\documentclass[]{interact}
\usepackage{amssymb}
\usepackage{amsmath}
\usepackage{hyperref}
\usepackage{hyperref}
\usepackage{comment}
\hypersetup{
    colorlinks=true,
    linkcolor=cyan,
    filecolor=cyan,
    urlcolor=cyan,
    citecolor=cyan,
}
\usepackage{natbib}
\usepackage{url}
\usepackage{epstopdf}
\usepackage{subfigure}
\usepackage{multirow}
\usepackage{natbib}
\bibpunct[, ]{(}{)}{,}{a}{}{,}
\renewcommand\bibfont{\fontsize{10}{12}\selectfont}

\theoremstyle{plain}
\newtheorem{theorem}{Theorem}[section]

\usepackage{CJKutf8}

\theoremstyle{definition}
\newtheorem{definition}[theorem]{Definition}

\usepackage{algorithm}
\usepackage{algorithmic}
\usepackage{amsfonts}
\usepackage{amsmath}
\usepackage{multirow}
\usepackage{tabularx}   
\usepackage{xcolor}
\usepackage{pifont}
\theoremstyle{remark}

\begin{document}


\title{Automated Quality Assessment of Geospatial Vector Data: A GeoAI Approach using Spatial Representation Learning}

\author{
\name{
Hao Li\textsuperscript{a},
Chen Chu\textsuperscript{b},
Filip Biljecki\textsuperscript{c,d},
Cyrus Shahabi\textsuperscript{b},
and Wenwen Li\textsuperscript{e}\thanks{CONTACT Hao Li. Email: hao.li@nus.edu.sg}}
\affil{
\textsuperscript{a} Department of Geography, National University of Singapore, Singapore; 
\textsuperscript{b}University of Southern California;
\textsuperscript{c}Department of Architecture, National University of Singapore, Singapore; 
\textsuperscript{d}Department of Real Estate, National University of Singapore, Singapore; 
\textsuperscript{e}School of Geographical Sciences and Urban Planning, Arizona State University, USA; }
}

\maketitle

\begin{abstract}

Geospatial vector data quality is a foundational research topic in GIS, yet classic rule-based quality assessment algorithms often struggle with diverse urban morphologies and massive data volumes. Recently, Geospatial Artificial Intelligence (GeoAI) shows promising potential for automating geospatial analysis, while its application to native vector data remains largely underexplored. To fill this research gap, we proposed Topo4Vec, an automated GeoAI framework, designed for scalable vector data quality assessment via advanced Spatial Representation Learning (SRL). Specifically, Topo4Vec relax the labor-intensive manual annotation process via topological error simulation, such as overlapping polygons and street network connectivity errors e.g., overshoots and undershoots. Then, it leverages state-of-the-art SRL approaches to encode complex, native vector geometries (e.g., polylines and polygons) into a latent space where topological errors are isolated from valid ones. A systematic performance evaluation across three study areas (Los Angeles, Munich, and Singapore) demonstrates the effectiveness and robustness of Topo4Vec, achieving a peak accuracy of 0.99 for detecting overlapping building footprints and 0.60 for overshoots and undershoots in street networks. Moreover, lessons learned from Topo4Vec shed a promising light into a scalable and autonomous GeoAI approach for large-scale vector data consistency and quality monitoring within the fast-growing geospatial data ecosystems. The code and data used in the paper are made openly available in \url{https://figshare.com/s/612148eeb4bccadbd715}.
    
\end{abstract}


\section{Introduction}  

High-quality geospatial vector data serves as the foundational basis for modern Geographic Information Systems (GIS) \citep{egenhofer1993critical, duckham2000assessment}, facilitating critical applications ranging from urban planning \citep{chen2006automatically} to disaster response \citep{de2015geographic} and many more. The spatial accuracy, consistency, and topological integrity of the data directly dominate the reliability of downstream spatial analyses and real-world decision-making processes \citep{barron2014comprehensive, liu2016rethinking}. As vector datasets continue to grow exponentially through automated mapping and crowdsourcing, identifying and rectifying data quality issues has become an increasingly sophisticated challenge \citep{haklay2010how, fan2014quality}. Therefore, establishing robust, automated quality assessment methods is key to ensure the trustworthiness and operational utility of spatial data ecosystems \citep{goodchild2013quality}. However, traditional rule-based algorithms (e.g., the Dimensionally Extended nine-Intersection Model (DE-9IM)~\citep{clementini1993small}) have long been the standard practice for quality assurance, they often struggle with the complex, massive appearances of errors found in dense urban environments and become computationally expensive when scaling up. Even the same types of overlapped buildings or overshoot streets, they can look very different cross cities. In this context, it is an important task for modern GIS systems to develop automatic and accurate geospatial vector data quality assessment approach.

Recently, Geospatial Artificial Intelligence (GeoAI) has achieved great success in advancing spatial data analysis and their GIS applications \citep{li2022geoai, liu2022review}, though its developments have been disproportionately focused on mainly raster data such as satellite imagery or street-view imagery \citep{hu2024five, li2025cross}. The application of GeoAI directly to native vector data with complex, non-Euclidean geometries like polygons and polylines remains comparatively underexplored \citep{mai2024srl}. Since vector data lacks the regular grid structure of pixels, classic learning schema the intricate topological relationships and geometric dependencies required for quality assurance becomes less effective \citep{chu2019geo, mai2022review}. As a consequence, there is a pressing research need in developing dedicated, native, and vector-based GeoAI methodologies capable of automatically assessing topological integrity without relying on intermediate rasterization steps.

To bridge the gap between complex vector geometries and GeoAI models, spatial representation learning (SRL) has emerged as a promising approach to embed spatial entities into low-dimensional, dense embeddings while preserving their geometric and relational properties \citep{mai2024srl, liu2025representation}. While these neural representations have been successfully applied to a range of downstream tasks such as land-use classification \citep{liu2020dynamic}, trajectory prediction \citep{hu2023recognizing}, and point-of-interest recommendation \citep{huang2022estimating}, the application of SRL to automatic vector data quality assessment remains still very limited. Existing SRL approaches, such as \citet{wang2024random} and \citet{siampou2025poly2vec}, are largely designed to capture specific structural patterns or semantic similarities, rendering them largely insensitive to the highly localized, micro-scale geometric anomalies that characterize topological errors in vector data (e.g., a slight polygon intersection or a minute dangling node). In this context, the second critical research gap exists in developing SRL frameworks that are explicitly optimized to isolate and distinguish topological errors from valid, clean spatial structures.

The emerging paradigm of Autonomous GIS mandates that rigorous data quality checks serve as a foundational first step of any automated GIS workflow \citep{li2023autonomous}, which has been the case for long in spatial analysis. The implications of this prerequisite are clear and straightforward: in autonomous systems, such as self-driving vehicle navigation or dynamic digital twins, unverified topological errors can cascade exponentially, leading to critical navigational failures or flawed spatial reasoning. However, to seamlessly integrate learning-based quality assurance into these autonomous pipelines is considered as a key technical barrier due to a heavy reliance on human-curated training data \citep{barron2014comprehensive}. As GeoAI models require massive datasets containing explicitly annotated spatial errors, but these errors are inherently rare and labor-intensive to identify in map production environments, so the requirement for manual labeling contradicts the very premise of autonomy \citep{barron2014comprehensive, zhang2022deep}. Therefore, the third research gap refers to the absence of an automated, end-to-end workflow capable of simulating its own realistic topological errors to train robust quality assessment models without relying on exhaustive manual labeling.

\begin{figure}
    \centering
    \includegraphics[width=\linewidth]{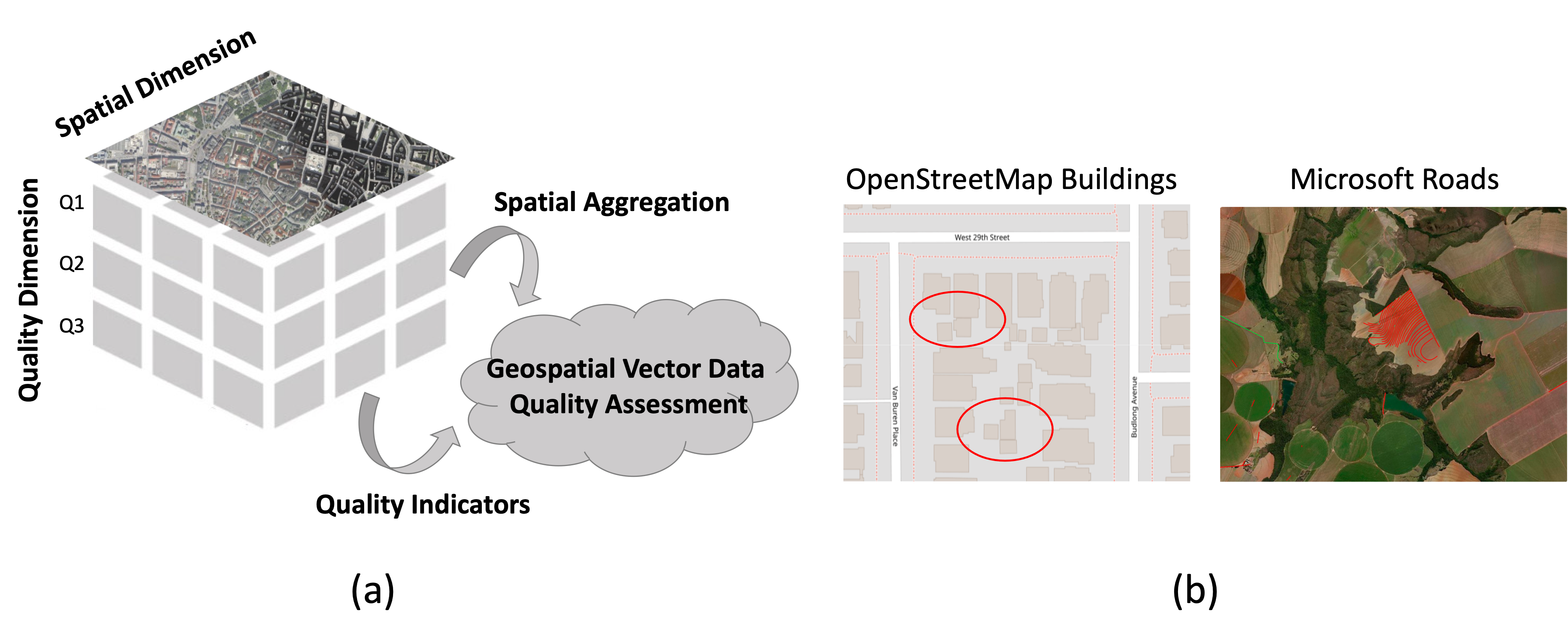}
    \caption{ A Diagram of Topo4Vec: (a) an Automated GeoAI-based framework for Geospatial Vector Data Quality Assessment, (b) real-world examples of vector data quality issues such as overlapping polygons in OpenStreetMap and inconsistent streets in Microsoft Roads.}
    \label{fig:diagram}
\end{figure}

To fill the identified research gaps, we presented \textbf{Topo4Vec}, an automated GeoAI-based framework for the topological quality assessment of vector data via SRL. Specifically, Topo4Vec can largely reduce the need of human annotations by designing a topological error simulation module that can simulate geometric anomalies with real-world vector datasets. Building upon this topological error simulation module, Topo4Vec then leverages state-of-the-art SRL approaches by applying contrastive learning to map complex, non-Euclidean vector geometries into a latent space where fine-scale topological errors are explicitly distinguished from valid spatial structures. Herein, we put focus on two most common vector data quality issues: (1) overlapping or duplicated building footprints, and (2) road connectivity errors, including overshoots and undershoots. By systematically evaluating Topo4Vec across the diverse urban morphologies of Singapore, Los Angeles, and Munich, this paper seeks to answer the following three main Research Questions (RQs):
\begin{itemize}
    \item \textbf{RQ1:} How can SRL and contrastive learning be synergistically combined to effectively capture and distinguish fine-scale topological errors in native vector data format?
    \item \textbf{RQ2:} To what extent can an automated topological error simulation workflow substitute for labor-intensive manual annotations to train robust GeoAI quality assessment models?
    \item \textbf{RQ3:} How does the model performance and generalizability vary when transfers across distinct urban morphologies, such as Singapore, Los Angeles, and Munich?
\end{itemize}

The remainder of this paper is organized as follows: Section 2 reviews related work on vector data quality assessment and spatial representation learning. Section 3 elaborates on the proposed Topo4Vec method together with the topological error simulation approaches. Section 4 presents experimental results on three cities. Section 5 discusses key findings as well as limitations and future directions, and Section 6 concludes the paper by highlighting the main contributions.

\section{Related Work}

\subsection{Geospatial Vector Data Quality Assessment}

The assessment of geospatial vector data quality has been a long-standing task of GIS and spatial data science. Historically, vector data quality assessment frameworks are broadly categorized into extrinsic and intrinsic approaches~\citep{goodchild2007citizens, senaratne2017review}. Extrinsic approaches, or reference-based methods, evaluate spatial data by comparing it against an authoritative ``ground truth'' dataset~\citep{haklay2010how}. While effective for assessing absolute positional and thematic accuracy, extrinsic methods are heavily constrained by the availability, high costs, and temporal frequency of the reference data. In contrast, intrinsic approaches are reference-free, assessing quality based on the internal characteristics of the dataset itself~\citep{barron2014comprehensive}. Particularly, for crowdsourced data like OpenStreetMap (OSM), intrinsic quality indicators are developed by analyzing user editing pattern, contributor history, and logical consistency to estimate data reliability without external ground truth dataset \citep{li2020exploration, li2022improving, herfort2023spatio}.

A fundamental spatial theory of quality assessment is the rule-based evaluation of topological consistency~\citep{egenhofer1991point}. Grounded in topological frameworks such as the Dimensionally Extended nine-Intersection Model (DE-9IM)~\citep{clementini1993small}, rule-based approaches enforce explicit geometric checks to identify inconsistencies and anomalies. Existing research has also explore the usage of different rule-based approaches in assessing the data quality of vector data, such as building polygons \citep{fan2014quality}, street networks \citep{, brovelli2017towards}, and attributes \citep{ , biljecki2023quality}. Based on explicit spatial rules, these quality assessment approaches can deterministically detect topological errors such as overlapping building polygons, slivers, and dangling street network nodes (e.g., overshoots and undershoots).

Despite the effectiveness, classic rule-based approaches show significant limitations when applied to modern, massive-scale vector datasets. Specifically, they suffer from inherent rigidity and high computational overhead~\citep{girres2010quality}. Rule-based approaches require domain experts to exhaustively define, threshold, and maintain topological constraint logic. Furthermore, the pre-defined rules may lack contextual awareness, as they frequently struggle to differentiate between genuine topological errors and complex, valid spatial context (e.g., multi-level infrastructure, complex interchanges) without triggering massive amount of false positives. In this context, a significant research gap exists for automated vector data quality assessment approaches. There is a pressing need for methodologies that can automatically learn the underlying topological semantics and contextual nuances of vector data to enable scalable, autonomous quality assessment \citep{li2016geospatial}.

\subsection{Neural Representation of Geospatial Entities}

The GeoAI community is actively moving away from rigid geometric rules and manual feature engineering. Instead, researchers are increasingly adopting Spatial Representation Learning (SRL) to encode points, lines, and polygons into general-purpose neural representations \citep{mai2024srl, liu2025representation}. At its core, SRL takes complex, non-Euclidean vector data and maps it into dense, low-dimensional embeddings. The key advantage is that it preserves the structural, spatial, and topological properties of the original geometries while benefit from its learning-based nature to accommodate the large amount of data and real-world complexities. Currently, the literature tackles these geometric challenges through three main approaches.

The first approach focuses on modeling spatial relationships using Graph Neural Networks (GNNs)~\citep{liu2025graph}. Since geographic features like street networks naturally form graphs—where intersections act as nodes and the streets themselves are edges, therefore GNNs are a logical choice for pooling neighborhood topology~\citep{Yan2021}. These models excel at capturing how linear features connect and depend on one another \citep{wu2020comprehensive}. However, traditional GNNs are designed for discrete graph structures. Because of this, they usually struggle to handle the continuous coordinate variations found in complex polygonal shapes, such as building footprints~\citep{polygnn}.

A second line of research uses spectral analysis and Fast Fourier Transforms (FFT)~\citep{mai2023towards} to better handle the complexities of polylines and polygons. Poly2Vec~\citep{siampou2025poly2vec} is a recent example, offering an FFT-based SRL framework capable of handling points, lines, and polygons all at once. Previous methods often relied on Recurrent Neural Networks (RNNs) to process boundary coordinates sequentially, which made them overly sensitive to the starting vertex and the density of sampled points. Poly2Vec bypasses this issue by applying FFT directly to the coordinate sequences of the boundaries. This shifts the variable-length spatial data into a fixed-length frequency representation. By focusing on the dominant frequencies, the model creates embeddings that remain invariant to both translation and rotation, capturing the true morphological shape of the vector data.

Finally, a third approach seeks to build unified spatial representations, for instance in Geo2Vec~\citep{geo2vec_paper}. Geo2Vec introduces a generalized architecture that places various geographic entities into a shared, distance-aware latent space. While the authors originally designed it for downstream tasks like semantic classification and spatial retrieval, the framework shows obvious potential for automating vector data quality assessment. By extending these advanced SRL models, one can explicitly map complex topological relationships straight into a latent space. There, micro-scale geometric errors—like overlapping buildings or disconnected streets—could be easily separated from valid geometries. Recognizing this potential was the primary motivation for our work, serving as the theoretical basis for Topo4Vec.

\subsection{Geographical Generalizability of GeoAI}

Geographical generalizability, refers to the capability of a GeoAI model to replicate or generalize the model's prediction ability across space \citep{li2023rethink}, which is preliminarily coined as \textbf{replicability across space} \cite{goodchild2021replication}. Geographical generalizability is a well-known challenge for GeoAI. In the remote sensing domain, this issue has been extensively studied \citep{xu2023universal}. Researchers actively employ domain adaptation and transfer learning techniques to mitigate spatial domain shifts in raster imagery caused by varying sensors, atmospheric conditions, or regional visual characteristics~\citep{tuia2016domain}. These domain adaption methods successfully align feature distributions between source and target geographic domains, enabling models to adapt to new environments without retraining from scratch \citep{hong2023cross}.

However, the geographical generalizability of GeoAI models operating directly on native vector data is surprisingly underexplored. Vector data encodes explicit geometric and topological structures, such as street network layouts and building arrangements, which all vary significantly across different urban morphologies \citep{biljecki2022global}. For instance, the highly planned, grid-based layout of a modern North American city presents a fundamentally different structural domain than the irregular, organic layout of a historical European city center. Structural domain shifts in non-Euclidean vector spaces present a fundamentally challenge than visual domain shifts in pixel grids. Therefore, how to enable vector-based representation approaches better adapt and generalize across diverse global urban morphologies remains a critical research gap, which largely motivates our work here.

\section{Method}
\begin{figure}[!t]
    \centering
    \includegraphics[width=0.9\linewidth]{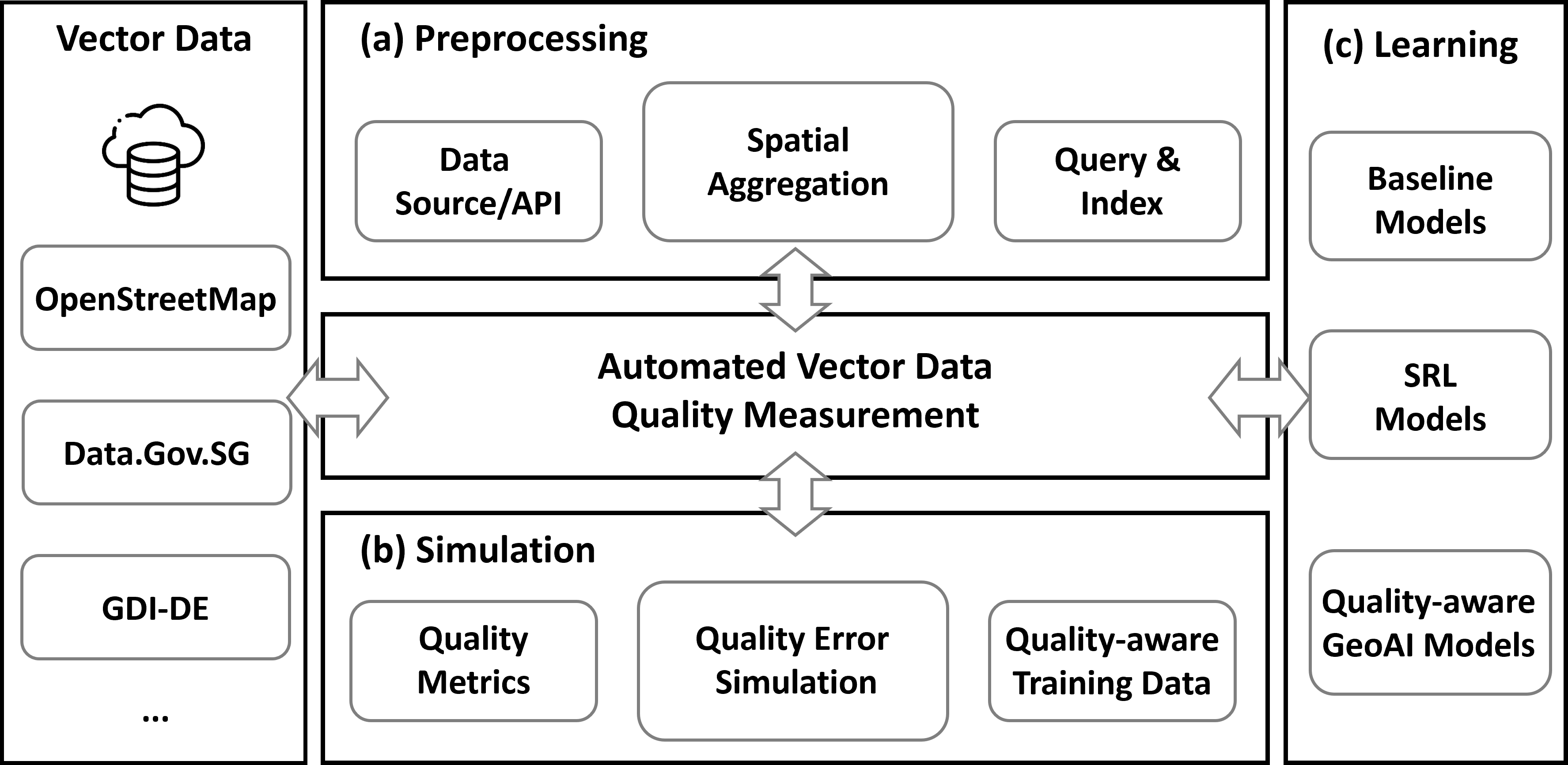}
    \caption{The technical workflow of Topo4Vec, detailing the data pipeline through three primary phases: (a) data preprocessing, (b) topological error simulation, and (c) neural representation learning for automated vector data quality measurement.}
    \label{fig:workflow}
\end{figure}

\subsection{Preliminaries}

\begin{definition}[Geospatial Vector Entities]
A geospatial vector entity $v \in \mathbb{R}^2$ can be represented by a list $P \in \mathbb{R}^{N \times 2}$, where each item is a point $(x, y)$, and $N$ is the total number of points. Classic vector data consists of three polymorphic types, namely point, polyline, and polygon, which are distinguished by the organization and relationships among these points.
\end{definition}

\begin{definition} [Spatial Representation Learning]
Given a dataset of geospatial vector entities \( D = \{v\} \in \mathbb{R}^{N \times 2} \), the goal of SRL is to define an non-linear encoding function \( E_{\theta}(v): \mathbb{R}^{N \times 2} \rightarrow \mathbb{R}^d \), parameterized by \( \theta \), that maps each geometry \( v \) to a \( d \)-dimensional 1D vector \(e_\theta\), namely embeddings. 
\end{definition}

\begin{definition}[Common Topological Errors]
Given a dataset $D$ of geospatial entities, a topological error denotes a violation of pre-defined spatial and topological relationship. Herein, we mainly focused on two primary errors using indicator functions $\mathbb{I}[\cdot]$:

\textbf{Overlapping Polygons:} For two polygons $p_i, p_j \in D$ with interiors $Int(v)$, an overlap occurs if their intersection yields a positive area than a strict tolerance $\epsilon > 0$: 
    \begin{equation}
        E_{overlap}(p_i, p_j) = \mathbb{I}\big[\mu\big(Int(p_i) \cap Int(p_j)\big) > \epsilon\big]
    \end{equation} 
    
\textbf{Over/Under-shoots Polylines:} For a connected network of polylines $L \subset D$, let $p$ denote the terminal nodes of $v_i$, and $d(p, v_j)$ the shortest Euclidean distance from node $p$ to polyline $v_j$. An error occurs if $p$ falls within a strict tolerance $\epsilon > 0$ without achieving exact topological connectivity (where $d=0$):
    \begin{equation}
         E_{shoot}(v_i, v_j) = \mathbb{I}\Big[0 < \min_{p \in v_i} d(p, v_j) \leq \epsilon\Big] 
    \end{equation}

Details about the topological error simulation will be provided in the following section.
\end{definition}

\subsection{Topological Error Simulation}

As shown in Figure \ref{fig:diagram}, an essential step of Topo4Vec is to simulate different types and magnitude of vector data quality errors for each study areas, which are then used an training data for the quality assessment models \citep{egenhofer1993critical}. Herein, following the aforementioned definition of common topological errors, we elaborated on the design of our topological error simulation workflow.


\begin{figure}[!t]
    \centering
    \includegraphics[width=0.8\linewidth]{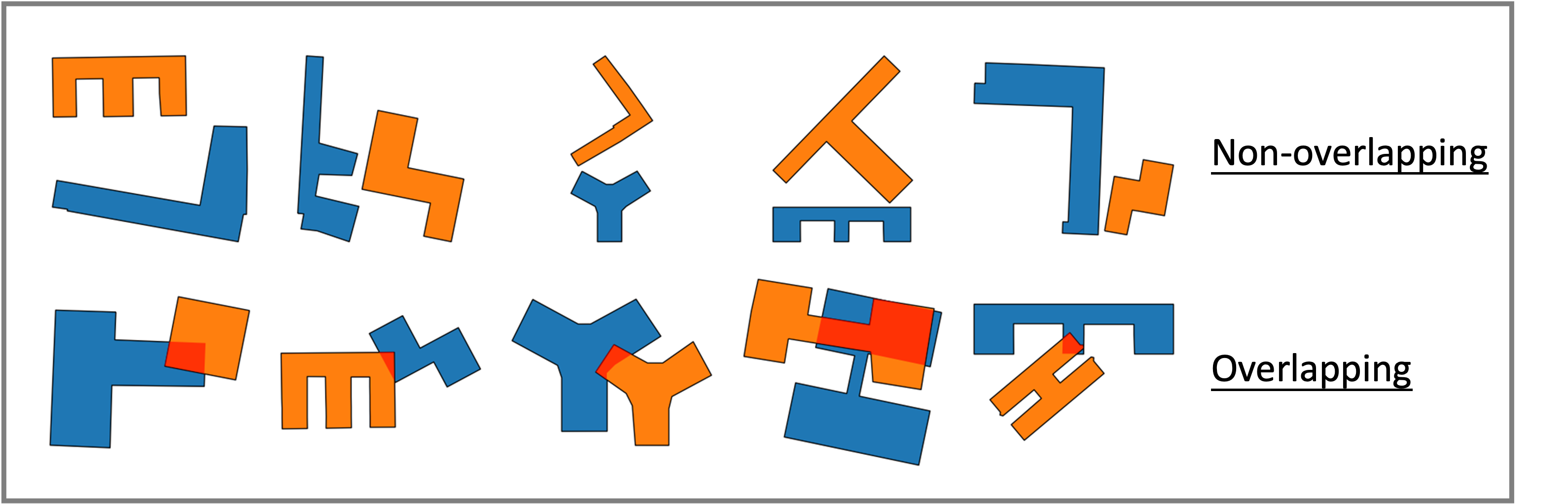}
    \caption{Visual examples of simulated topological errors in building footprints, contrasting valid non-overlapping structures with simulated overlapping polygons.}
    \label{fig:topo_error_polygon}
\end{figure}

\textbf{Overlapping Polygons Simulation:} To ensure the generalization of the learned spatial representation function $E_\theta$, we designed a stochastic simulation pipeline that simulate a pair of random but distinct polygons $(p_i, p_j)$ into a labeled training sample. Let $G$ be the set of building footprints from a specific urban cities.

\textbf{Geometric Normalization and Augmentation:} For each polygon $v \in G$, we fist applied a normalization function $T(v) = N(v)$ to noramlize the vertex coordinates $p = \{(x, y)\}$ to a unit square $[0,1] \times [0,1]$ , which helps to achieve scale invariance in the simulation.

\textbf{Rotation Invariant Simulation:} To ensure the embedding is rotation-invariant, we then applied a random rotation matrix $R_\alpha$ with a stochastic angle $\alpha \in [0, 2\pi)$:
    \begin{equation}
        \phi(p) = \{ R_\alpha p \mid p \in \psi(p) \}
    \end{equation}

\textbf{Stochastic Overlapping Simulation:} Let $B(v)$ denote the Minimum Bounding Rectangle of polygon $p$. To simulate random overlaps, we defined a translation vector $\vec{\delta}$ that forces an intersection between two randomly selected polygons $p_i, p_j \in G$. We defined the overlap proportion $\rho$ as the ratio of the intersection area to the area of the smaller polygon:
\begin{equation}
    \rho(p_i, p_j) = \frac{\mu\big(\text{Int}(p_i) \cap \text{Int}(p_j)\big)}{\min(\mu(p_i), \mu(p_j))}
\end{equation}
Where the simulation process samples a target proportion $\rho^* \sim \mathcal{U}(\rho_{\min}, \rho_{\max})$ from a Gaussian normal distribution to ensure the randomness in the simulation results.

As a result, we have the following simulation set $E_{overlap}(p_i, p_j)$, where a strict non-overlapping rule checks their interiors, $Int(v)$, with no overlapping space, and an polygon overlapping error occurs when the area of their intersection is a positive value. To reduce bias toward classification, we constructed the training set $E_{overlap}(p_i, p_j)$ as a balanced collection of positive (e.g., Overlapped) and negative (e.g., Non-overlapped) samples.

\begin{equation}
    E_{overlap}(p_i, p_j) = 
    \begin{cases} 
    1, & \text{if } \big(Int(p_i) \cap Int(p_j)\big) > 0 \\
    0, & \text{otherwise}
    \end{cases}
\end{equation}

\textbf{Undershoots and Overshoots Street Network Simulation:} 


\begin{figure}
    \centering
    \includegraphics[width=0.8\linewidth]{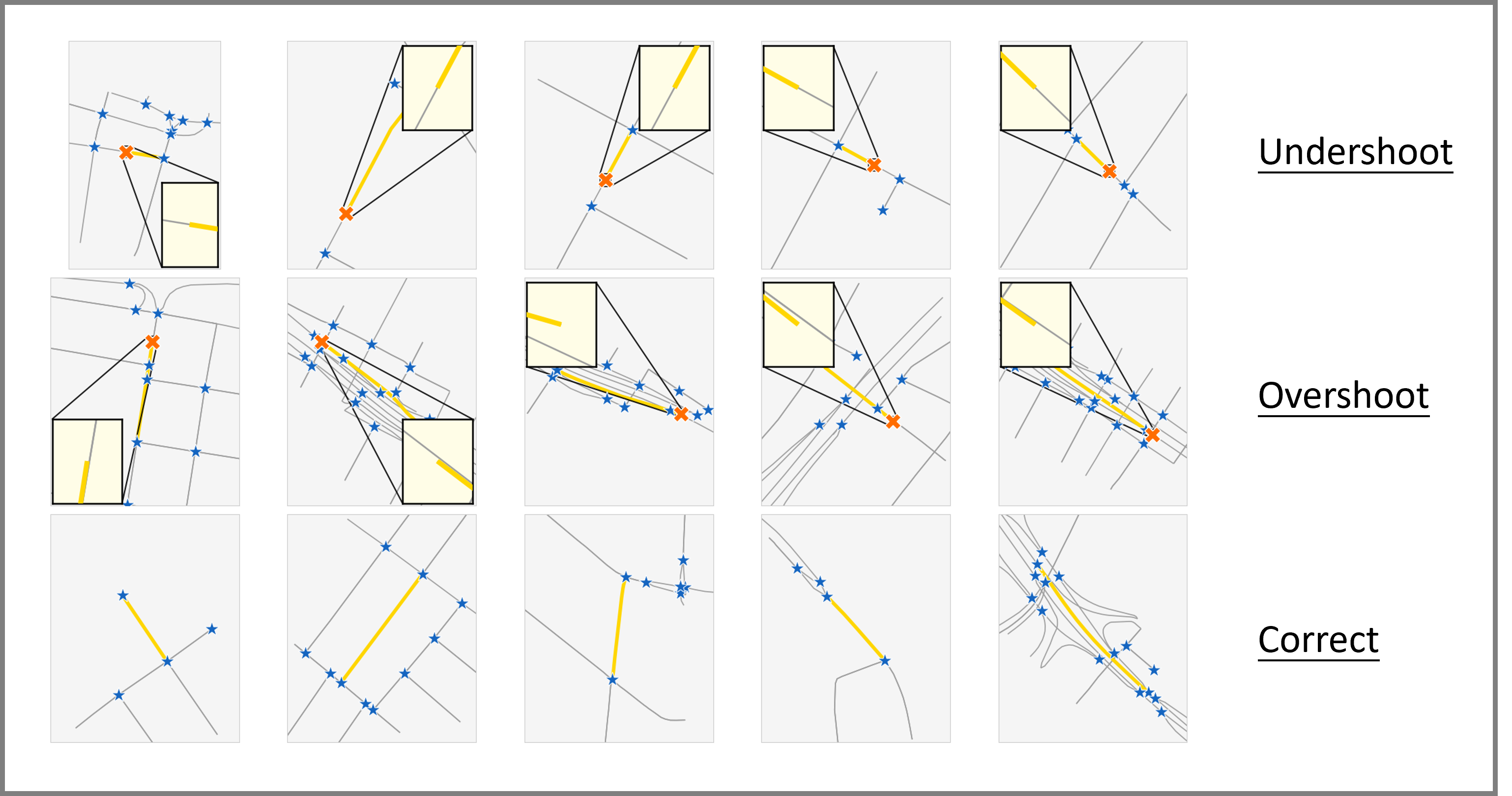}
    \caption{Visual examples of simulated connectivity anomalies in street networks, distinguishing between simulated undershoot errors, overshoot errors, and correct polyline topologies}
    \label{fig:topo_error_polyline}
\end{figure}

Given a set of polyline entities $L$ (e.g., a street network). For a polyline $v_i \in L$, let $n \in \partial v_i$ denote a terminal node. Let $v_j$ be the targeted adjacent polyline, where $d(n, v_j) = \min_{m \in v_j} \|n - m\|_2$ is the shortest Euclidean distance, where $m \in v_j$ refer to other nodes. Herein, a connectivity error occurs when $d(n, v_j)$ falls within a strict snap-tolerance $\epsilon > 0$ without forming an exact node (i.e., $d=0$). We distinguished these cases based on intersection.

Specifically, the generation of training data for under- and overshoots street networks is designed as a stochastic transformation of a simplified 1-complex within a constrained urban graph. 

\textbf{Topological Junction Selection}
Let $\mathcal{J} \subset L$ denote the set of junction nodes where $\text{deg}(v) \geq 3$. For each $j \in \mathcal{J}$, the incident edge set is defined by $St(j) = \{e \in E \mid j \in \partial e\}$. The simulation breaks these junctions as potential locations for connectivity errors.

\textbf{Constrained Geometric Perturbation}
Topological errors are introduced via a scaling operator applied to the terminal nodes of a selected edge $e_i \in St(j)$. We enforced a \textit{Mutual Exclusion Constraint} to maintain a stable local topology, ensuring only one incident edge per junction is being perturbed. The corrupted terminal node  $n'_i$ is computed as:
\begin{equation}
    n'_i = n_{i-1} + (1 + \lambda) (n_i - n_{i-1})
\end{equation}
where $\lambda$ is a stochastic scalar sampled to produce distinct error classes. Herein,
Overshoot (i.e., $\mathcal{E}_{over}$) refers to $\lambda > 0$, where the polyline extends beyond the junction node, and Undershoot (i.e., $\mathcal{E}_{under}$) refers to $\lambda < 0$, creating a gap within the snapping threshold.

To facilitate scalable simulation, the augmented dataset is partitioned into $K$ disjoin geographical subsets $\{D_1, \dots, D_K\}$. We defined a feature inheritance function $\mathcal{I}$ that preserves the semantic attribute vectors $\mathbf{a} \in \mathbb{R}^m$ from the original polylines, ensuring that the SRL function $E_\theta$ receives consistent non-spatial cues across both valid and corrupted samples.

\textbf{Topological Integrity Postprocessing}
An additional post-processing step ensures the topological integrity of each training subsets. For every corrupted edge $e' \in D_k$, we enforced a closure property:
\begin{equation}
    \forall e' \in D_k, \quad \{e_j \mid \text{ID}(e_j) \in \text{Adj}(e')\} \subseteq D_k
\end{equation}
Where $\text{Adj}(e')$ denotes the set of topological neighbors. If a neighbor is missing due to partitioning, it is included again to preserve the local graph manifold required for the data quality assessment tasks.

As a result, we have the following two sets of topological errors for polylines, where:

\textbf{Undershoot Errors:} The polyline $v_i$ falls short of $v_j$. Geometrically, the geometries do not intersect ($v_i \cap v_j = \emptyset$), but the endpoint is within the tolerance threshold:
\begin{equation}
    E_{under}(n) = 
    \begin{cases} 
    1, & \text{if } 0 < d(n, v_j) \leq \epsilon \text{ and } v_i \cap v_j = \emptyset \\
    0, & \text{otherwise}
    \end{cases}
\end{equation}
    
\textbf{Overshoot Errors:} The polyline $v_i$ connected with $v_j$ but erroneously extends past the intersection point, leaving a dangling segment. Geometrically, two polylines intersect ($v_i \cap v_j \neq \emptyset$), but the endpoint $p$ extends beyond the intersection by a distance within the tolerance:
\begin{equation}
    E_{over}(n) = 
    \begin{cases} 
    1, & \text{if } 0 < d(n, v_j) \leq \epsilon \text{ and } v_i \cap v_j \neq \emptyset \\
    0, & \text{otherwise}
    \end{cases}
\end{equation}

\subsection{Representation Learning of Geospatial Vector Entities}

To learn a robust neural representation of polygons and polylines, we employed the Geo2Vec model \citep{geo2vec_paper} as the state-of-the-art SRL approach. Herein, a key reason is that Geo2Vec's Signed Distance Field (SDF) approach offers an explicit and intuitive representation capability of shape, distance, and topological relationships, which is an unique advantage over many of its FFT and GNN peer approaches \citep{oleynikova2016signed}. In the rest of this section, we will elaborate on the details of SDF approach and the way how it is used in Topo4Vec.


\begin{figure}[!t]
    \centering
    \includegraphics[width=0.8\linewidth]{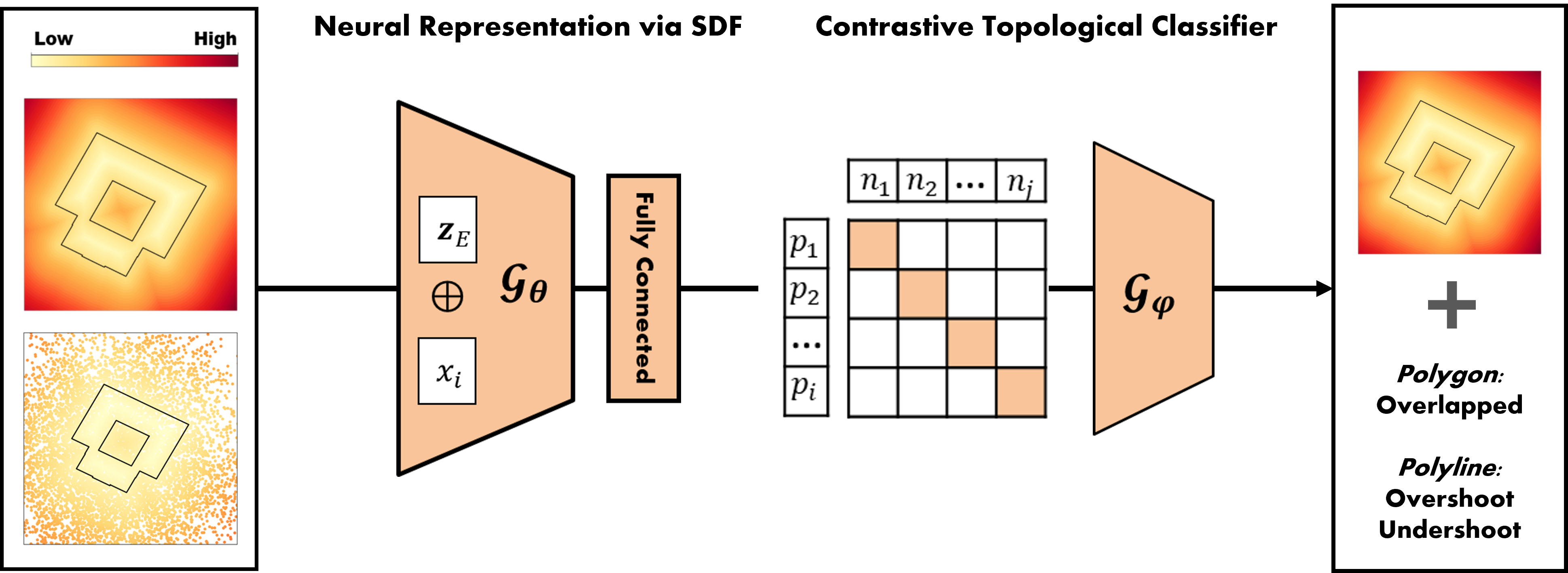}
    \caption{The architecture of the Geo2Vec model used in the Topo4Vec, highlighting the neural representation pipeline via Signed Distance Functions (SDF) and the contrastive topological classifier designed to detect overlapping polygons and over- and undershoot street networks.}
    \label{fig:topovec_archi}
\end{figure}

\textbf{Definition and SDF Construction:} For a vector entity $E$ (i.e., polygon or polyline) and a query coordinate $\mathbf{x} \in \mathbb{R}^2$, the Signed Distance Function $\mathbf{SDF}(\mathbf{x}, E)$ returns the shortest scalar distance $s$ to the boundary $\partial E$. For any entity with a defined spatial extent, the sign distinguishes the interior from the exterior:
\begin{equation}
    \mathbf{SDF}(\mathbf{x}, E) = 
    \begin{cases} 
    d(\mathbf{x}, \partial E) & \text{if } \mathbf{x} \in \text{Ext}(E) \\
    0 & \text{if } \mathbf{x} \in \partial E \\
    -d(\mathbf{x}, \partial E) & \text{if } \mathbf{x} \in \text{Int}(E)
    \end{cases}
\end{equation}

By defining the SDF as a scalar field over a continuous spatial domain $\Omega_E \subseteq \mathbb{R}^2$, we capture the global location and local shape morphology within a singular, differentiable format. We proposed a neural network $\mathcal{G}_{\theta}$ to approximate this field based on a set of sampled points $X_E$:
\begin{equation}
\mathcal{G}_{\theta}(\mathbf{x}_i, \mathbf{z}_E) \approx \mathbf{SDF}(\mathbf{x}_i, E), \quad \forall \mathbf{x}_i \in X_E \subset \Omega_E
\end{equation}

\textbf{Probabilistic Inference of Latent Embeddings:} Following \cite{park2019deepsdf} and \cite{geo2vec_paper}, this SDF reconstruction task can be formulated from a probabilistic perspective. We defined the posterior distribution over the latent code $\mathbf{z}_E$ given the sampled spatial observations $X_E$ as:
\begin{equation}
p_{\theta}(\mathbf{z}_E \mid X_E) = p(\mathbf{z}_E) \prod_{(\mathbf{x}_i, s_i) \in X_E} p_{\theta}(s_i \mid \mathbf{z}_E; \mathbf{x}_i)
\end{equation}
where $p(\mathbf{z}_E)$ denotes the prior distribution over latent codes, which we assume to follow a multivariate Gaussian $\mathcal{N}(\mathbf{0}, \sigma_z^2 \mathbf{I})$. 

Then, we can calculate the conditional likelihood $p_{\theta}(s_i \mid \mathbf{z}_E; \mathbf{x}_i)$ is defined such that it is inversely proportional to the reconstruction error:
\begin{equation}
p_{\theta}(s_i \mid \mathbf{z}_E; \mathbf{x}_i) \propto \exp\left(-\mathcal{L}(\bar{s}_i, s_i)\right)
\end{equation}
where $\bar{s}_i = \mathcal{G}_{\theta}(\mathbf{z}_E, \mathbf{x}_i)$ is the predicted signed distance at coordinate $\mathbf{x}_i$, and $\mathcal{L}$ is the SDF loss function. 

Consequently, maximizing the posterior probability is equivalent to minimizing the regularized loss function across the dataset $G$:
\begin{equation}
\mathcal{L}_{\text{SRL}} = \sum_{E \in G} \left( \sum_{(\mathbf{x}_i, s_i) \in X_E} \mathcal{L}(\mathcal{G}_\theta(\mathbf{z}_E, \mathbf{x}_i), s_i) + \frac{\gamma}{\sigma_z^2} \|\mathbf{z}_E\|_2^2 \right)
\end{equation}
Where the hyperparameter $\gamma$ governs the strength of the prior constraint. For purely positional representations, we set $\gamma = 0$ to prevent the collapse of the latent space, thereby preserving the natural diversity of global spatial distributions across geo-entities.

\textbf{Network Configuration and Optimization} \\
In fact, the backbone architecture of Geo2Vec is a deep Multi-Layer Perceptron (MLP) with LeakyReLU activation functions \citep{maas2013rectifier}. The choice of LeakyReLU is important as it makes sure that the network preserves non-zero gradients in the negative domain ($s < 0$), which then allows for precise supervision of the interior topological structures of vector geometries. Moreover, the input coordinate $\mathbf{x}_i$ is projected via positional encoding to capture high-frequency geometric details in the spectral domain, then concatenated with the latent representation $\mathbf{z}_E$. For more details about the positional encoding, one can refer to \cite{geo2vec_paper}.

\begin{algorithm}[!h]
    \caption{Geo2Vec Training Algorithm}
    \label{Learning}
    \renewcommand{\algorithmicrequire}{\textbf{Input:}}
    \renewcommand{\algorithmicensure}{\textbf{Output:}}

    \begin{algorithmic}[1]
        \REQUIRE Dataset $G=\{E\}$, Sample Density $\epsilon$, Axis-aligned Samples $N_{\text{axis}}$, Batch size $b$
        \ENSURE Optimized Latent Embeddings $\{\mathbf{z}_{E}\}_{E \in G}$
        \STATE Initialize $\mathcal{G}_{\theta}$ and latent codes $\{z_E\}_{E \in G} \sim \mathcal{N}(\mathbf{0}, \sigma_z^2 \mathbf{I})$
        \STATE Initialize $N_{\text{Edge}}$, $N_{\text{Vertex}}$, $\sigma$, and global sample pool $X_G \gets \emptyset$
        \FOR{each $E \in G$}
            \STATE $X_E \gets \text{Sample}(E, N_{\text{Edge}}, N_{\text{Vertex}}, N_{\text{axis}}, \sigma)$
            \STATE $X_G \gets X_G \cup \{(\mathbf{x}_i, s_i, E) \mid (\mathbf{x}_i, s_i) \in X_E\}$
        \ENDFOR
        \STATE Shuffle $X_G = \{(x_i, s_i, E_i)\}$
        \FOR{each mini-batch $\{(x_i, s_i, E_i)\}_{i=1}^{b} \subset X_G$}
            \STATE $\mathcal{L} = \mathcal{L}_{\text{Geo2Vec}}(\{(x_i, s_i, E_i)\}_{i=1}^b)$
            \STATE Update $\{\mathbf{z}_{E_i}\}_{E_i \in b}$ and $\mathcal{G}_{\theta}$ using $\mathcal{L}$ via backpropagation
        \ENDFOR
        \RETURN $\{\mathbf{z}_E\}_{E \in G}$
    \end{algorithmic}
\end{algorithm}

As shown in Algorithm \ref{Learning}, we jointly optimized the parameters $\theta$ and the latent codes $\{\mathbf{z}_E\}$ over large mini-batches for both polyline and polygon datasets. Finally, this combined vector $\mathbf{z}_E$ is then concatenated with the hidden states at each layer, serving as our SRL embeddings used for vector data quality assessment.

\subsection{Contrastive Learning of Topological Error}

In our setting, topological errors are generated through simulation, which provides an unique advantage: for each error samples, we can identify a highly similar non-error counterpart that differs only in fine-scale geometric details. These pairs then functions as hard negatives, where the distinction between valid and invalid topology is often ambiguous (e.g., micro-overshoots or near-touching geometries).

Under such conditions, conventional batch-wise supervised learning, which relies on global class labels and large sets of positives/negatives, may get confused, as many negative samples are trivially different. Instead, we adopt a pair-wise contrastive learning objective that explicitly focuses on these hard pairs, encouraging the model to learn fine-grained discriminative features.

Specifically, let $\mathbf{z}^+$ and $\mathbf{z}^-$ denote the embeddings of an error sample and its corresponding valid counterpart, respectively. A scoring function $s(\mathbf{z}) \in \mathbb{R}$ is learned to reflect the likelihood of topological error. We defined a margin-based contrastive loss function:

\begin{equation}
\mathcal{L}{\text{con}} = \mathbb{E}{(\mathbf{z}^+, \mathbf{z}^-)} \left[ \max\left(0, m - s(\mathbf{z}^+) + s(\mathbf{z}^-)\right) \right]
\end{equation}
where $m > 0$ is a margin hyperparameter. This objective enforces a relative ordering between paired samples, such that error samples receive higher scores than their hard negative counterparts by at least a margin $m$. From a contrastive perspective, this can be interpreted as pulling apart positive and negative pairs along a one-dimensional scoring axis, rather than within a normalized embedding space.

\textbf{Auxiliary Classification Objective}: To stabilize training and ensure that the learned scoring function is globally optimized, we introduced an auxiliary binary classification objective over both topological error and valid samples. 

Let $s(\mathbf{z}) \in \mathbb{R}$ denote the logit output of the scoring network, and $y \in {0,1}$ indicate whether the sample corresponds to a topological error. The binary cross-entropy loss is defined as:
\begin{equation}
\mathcal{L}{\text{bce}} = \mathbb{E} \left[ - , y \log \sigma\big(s(\mathbf{z})\big) - (1 - y)\log \big(1 - \sigma\big(s(\mathbf{z})\big)\big) \right]
\end{equation}

This objective encourages the MLP classifier to assign higher scores to error samples and lower scores to valid ones, complementing the pair-wise contrastive ranking by providing additional supervisions.

Finally, the overall training objective combines the pair-wise contrastive loss with the classification loss:

\begin{equation}
\mathcal{L}_{\text{Topo4Vec}} = \mathcal{L}{\text{con}} + \lambda  \mathcal{L}_{\text{bce}}
\end{equation}
where $\lambda$ balances the two components.


\subsection{Study area and Data}


\begin{figure}[!ht]
    \centering
    \includegraphics[width=\linewidth]{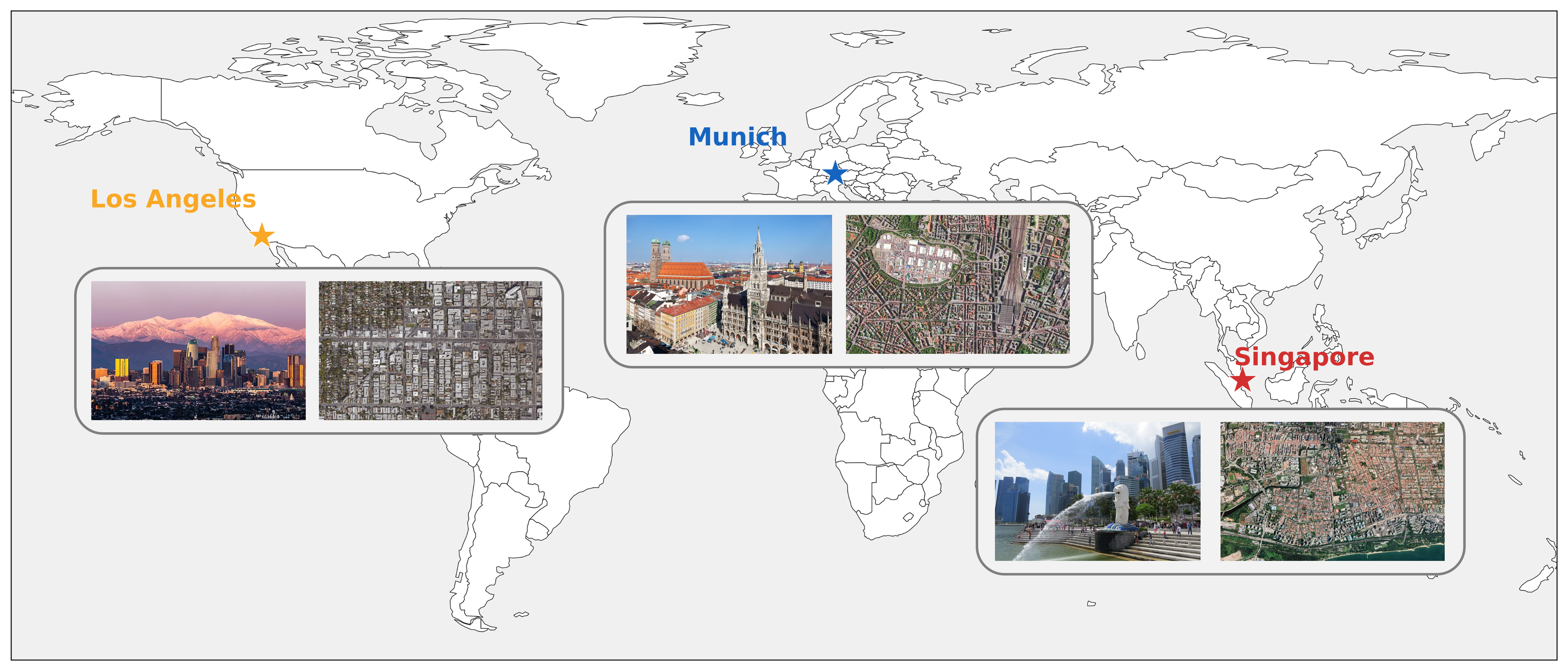}
    \caption{Three study area selected for evaluating the performance of Topo4Vec, namely Los Angeles (USA), Munich (Germany), and Singapore.}
    \label{fig:study_area}
\end{figure}

To evaluate the proposed Topo4Vec approach, three distinct metropolitan areas, specifically Los Angeles (USA), Munich (Germany), and Singapore, were selected as our study areas. Herein, for each cities, we considered 50,000 random buildings and a 500 $\times$ 500 meter split for the street network for each city. These three cities feature distinct urban morphologies, street network densities, and building typologies, providing a comprehensive and valid basis for assessing the performance of Topo4Vec across varying topological and geographical context.  

\begin{itemize}
    \item \textbf{Los Angeles (LA), USA:} Covering an area of approximately 1,302 km$^2$, Los Angeles is famous for its typical low-density, sprawling urban morphology as other North American cities. LA is characterized by a rectilinear street grid and predominantly orthogonal, single-family residential footprints, interspersed with high-density commercial areas. This topological pattern challenges the evaluation of the model on uniform, convex polygons and extensive polyline networks, where connectivity errors such as overshoots and undershoots are prevalent.
    
    \item \textbf{Munich, Germany:} With an administrative area of 310 km$^2$, Munich represents a traditional European city morphology characterized by mid-dense, well-planned development patterns. In Munich, building footprints usually feature complex, non-convex geometries, including contiguous block perimeters and internal courtyards. These spatial features present a higher degree of difficulty for Topo4Vec in terms of SRL and data quality assessment. Herein, the building footprint dataset for Munich is derived from the official Bavarian open data portal (\textit{Hausumringe})\footnote{\url{https://geodaten.bayern.de/opengeodata/OpenDataDetail.html?pn=hausumringe}}, maintained by the Bavarian State Office for Digitization, High-Speed Internet and Surveying.
    
    \item \textbf{Singapore:} Known as the small "Red Dot", Singapore covers roughly 734 km$^2$ with high-density, well-planned urban environment. Singapore's morphological structure is largely defined by public housing estates developed by the Housing \& Development Board (HDB). These estates consist of massive, repetitive geometric configurations, such as slab, point, and branched building footprints, that contrast significantly with European or US architectural styles. The explicit spatial constraints of HDB buildings provide an unique example for vector data SRL. Moreover, the HDB footprints were obtained from the official open dataset published via the Singapore Government Open Data Portal\footnote{\url{https://data.gov.sg}}.
\end{itemize}

\subsection{Evaluations}

In this section, we evaluated the effectiveness of Topo4Vec in vector data quality assessment (e.g., polygons and polylines). By formulating this as a classification task, we analyzed the performance of Topo4Vec across two types of common topological errors with both building polygons and street networks. Herein, the quantitative performance of all models is measured using a comprehensive suite of standard classification metrics: Precision, Recall, Accuracy, F1, and Area Under the Curve (AUC).

To examine the effectiveness and generalizability of Topo4Vec, we included several state-of-the-art SRL approaches with a fair ablation study. Specifically, for polygon SRL approaches, we compared NUFT \citep{mai2023towards}, Poly2Vec \citep{siampou2025poly2vec}, and Geo2Vec \citep{geo2vec_paper} for overlapping error of building footprints. For polyline SRL approaches, we compared JEPA \citep{li2024t}, MAE \citep{zhu2024controltraj}, Poly2Vec \citep{siampou2025poly2vec}, and Geo2Vec \citep{geo2vec_paper} for overshoot and undershoot errors of street networks.

\subsubsection{Topological Errors of Building Polygons}

Table \ref{tab:polygon_mlp_all} shows the experiment results of detecting overlapping polygons, featuring a clear performance difference across all three geographical regions. In this case, two key findings are observed: 

First, Geo2Vec is consistently outperforming both Poly2Vec and NUFT, resulting in the highest scores in almost every classification metric, achieving an almost perfect F1 of 0.99, an accuracy of 0.98, and a perfect AUC of 1.00 in Munich. Although NUFT establishes a solid baseline by maintaining F1 above 0.91 across all cities, the neural representations of Poly2Vec and Geo2Vec give a noticeable improvement, increasing the accuracy in Los Angeles and Singapore from 0.95 to 0.97 and 0.89 to 0.94, respectively.

Second, the classification results also highlight regional variations in the building complexity and model robustness. For example, all evaluated models experience a performance drop on the Singapore dataset compared to Los Angeles and Munich. For example, NUFT's accuracy drops to 0.89 and its F1 falls to 0.91 in Singapore, suggesting that the urban morphology pattern together simulated overlapping polygons in Singapore's unique urban environment are inherently more challenging. Despite the complexity, Geo2Vec demonstrates good robustness and effectiveness, which maintains a strong metric with F1 of 0.96 and an accuracy of 0.94 in Singapore, effectively mitigating the performance drop seen in the baseline models.

\begin{table*}[!htbp]
\centering
\small
\caption{Performances of \textbf{Overlapping Polygons Detection} in three study areas (with best result in bold, and all metrics in Mean $\pm$ Std, from 5 runs)}
\label{tab:polygon_mlp_all}
\resizebox{\textwidth}{!}{%
\begin{tabular}{@{} ll *{5}{c} @{}}
\toprule
City & Base & F1 & AUC & Accuracy & Precision & Recall \\
\midrule
\multirow{3}{*}{Los Angeles} & NUFT & $0.95 \pm 0.01$ & $0.95 \pm 0.03$ & $0.94 \pm 0.01$ & $0.95 \pm 0.01$ & $0.96 \pm 0.01$ \\
& Poly2Vec & $0.97 \pm 0.01$ & $\mathbf{0.99} \pm 0.00$ & $0.96 \pm 0.02$ & $0.97 \pm 0.01$ & $0.96 \pm 0.04$ \\
& Geo2Vec & $\mathbf{0.98} \pm 0.01$ & $0.99 \pm 0.01$ & $\mathbf{0.97} \pm 0.01$ & $\mathbf{0.99} \pm 0.00$ & $\mathbf{0.98} \pm 0.02$ \\
\midrule
\multirow{3}{*}{Munich} & NUFT & $0.96 \pm 0.00$ & $0.99 \pm 0.01$ & $0.96 \pm 0.00$ & $0.97 \pm 0.02$ & $0.96 \pm 0.01$ \\
& Poly2Vec & $0.98 \pm 0.01$ & $0.99 \pm 0.01$ & $\mathbf{0.98} \pm 0.01$ & $0.97 \pm 0.02$ & $\mathbf{0.99} \pm 0.01$ \\
& Geo2Vec & $\mathbf{0.99} \pm 0.01$ & $\mathbf{1.00} \pm 0.00$ & $\mathbf{0.98} \pm 0.02$ & $\mathbf{0.99} \pm 0.01$ & $\mathbf{0.99} \pm 0.02$ \\
\midrule
\multirow{3}{*}{Singapore} & NUFT & $0.91 \pm 0.02$ & $0.92 \pm 0.02$ & $0.89 \pm 0.02$ & $0.89 \pm 0.02$ & $0.93 \pm 0.02$ \\
& Poly2Vec & $0.94 \pm 0.01$ & $0.97 \pm 0.03$ & $0.93 \pm 0.02$ & $0.91 \pm 0.02$ & $0.96 \pm 0.02$ \\
& Geo2Vec & $\mathbf{0.96} \pm 0.02$ & $\mathbf{0.98} \pm 0.01$ & $\mathbf{0.94} \pm 0.02$ & $\mathbf{0.95} \pm 0.03$ & $\mathbf{0.97} \pm 0.01$ \\
\bottomrule
\end{tabular}%
}
\end{table*}

\begin{figure}[!htbp]
    \centering
    \includegraphics[width=0.9\linewidth]{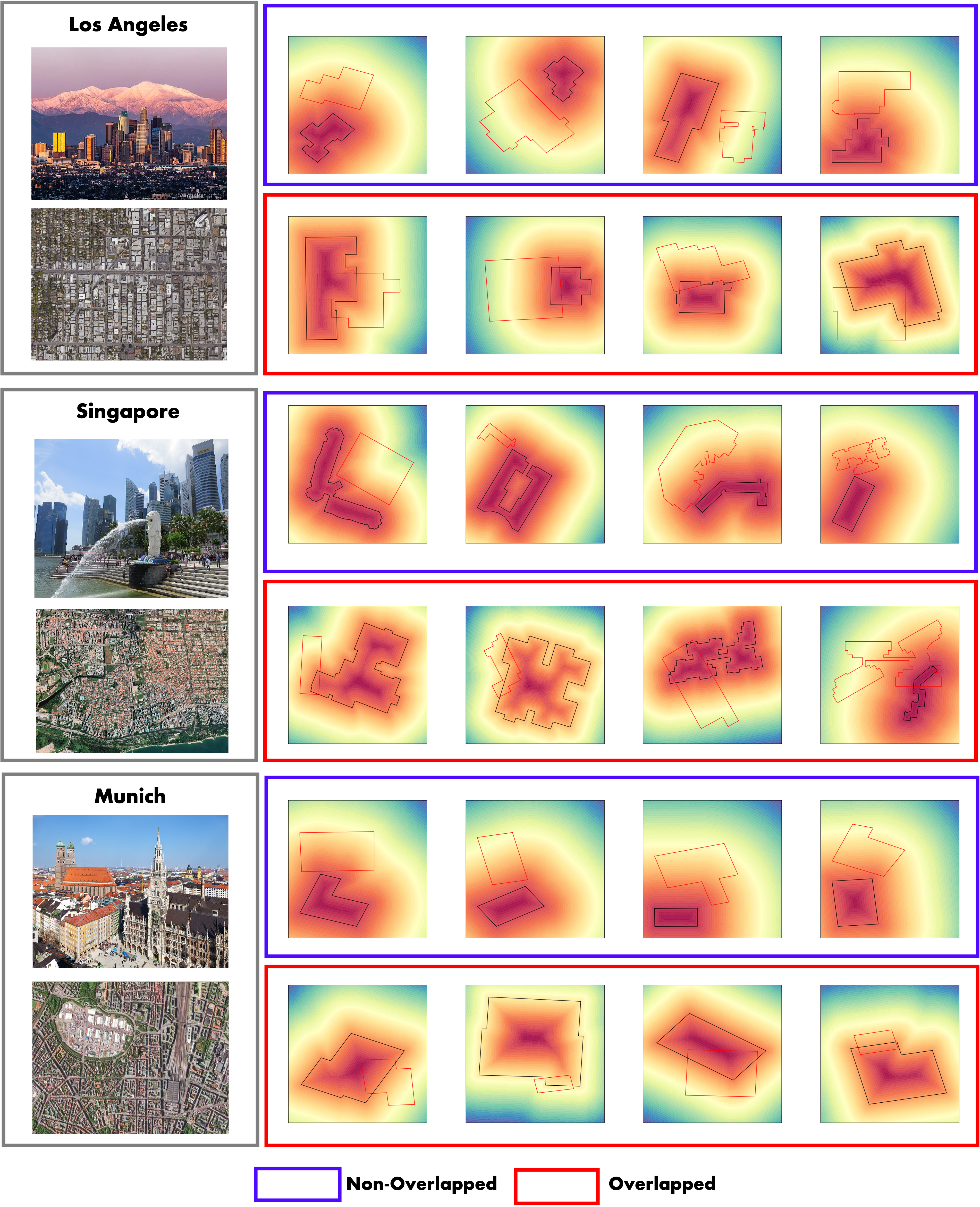}
    \caption{Selected visualizations of the learned SDF for non-overlapped and overlapped polygon pairs across Los Angeles, Singapore, and Munich.}
    \label{fig:polygon_examples}
\end{figure}

Moreover, a visual inspection of SDF maps learned by the Geo2Vec approach in Figure \ref{fig:polygon_examples}, w.r.t different cities (i.e., Los Angeles, Singapore, Munich) and different polygon pairs (i.e., non-overlapped and overlapped), further confirms these key findings. One can obviously see that building footprints in Singapore consist of more edges and complex structures, which is steered by the unique urban planning practices and socio-economic contexts. In this case, we argued that Geo2Vec not only yields higher accuracy, but also provides a more intuitive way of visualizing and understanding the neural representation that learned by the SRL models. This is considered as an essential advantage over classic GNNs or FFT-based SRL approaches.

\subsubsection{Topological Errors of Street Networks}

Moving on to topological errors of street networks, we considered mainly two common types of data quality issues, namely overshoots and undershoots. Table \ref{tab:overshoot_contrastive} and \ref{tab:undershoot_contrastive} presents the city-wise comparison of four different SRL approaches within the proposed Topo4Vec framework. Two key findings are observed in this case.

\begin{table*}[!htbp]
\centering
\small
\caption{Performances of \textbf{Overshoot Street Network} Detection in three study areas (with best result in bold, and all metrics in Mean $\pm$ Std, from 5 runs)}
\label{tab:overshoot_contrastive}
\resizebox{\textwidth}{!}{%
\begin{tabular}{@{} l l *{5}{c} @{}}
\toprule
City & Base & F1 & AUC & Accuracy & Precision & Recall \\
\midrule
\multirow{4}{*}{Los Angeles} 
  & JEPA & $0.60 \pm 0.01$ & $0.61 \pm 0.00$ & $0.58 \pm 0.00$ & $0.59 \pm 0.00$ & $0.61 \pm 0.02$ \\
  & MAE & $0.52 \pm 0.06$ & $0.59 \pm 0.00$ & $0.55 \pm 0.01$ & $0.57 \pm 0.00$ & $0.49 \pm 0.11$ \\
  & Poly2Vec & $\mathbf{0.64} \pm 0.00$ & $\mathbf{0.70} \pm 0.00$ & $\mathbf{0.64} \pm 0.00$ & $\mathbf{0.67} \pm 0.01$ & $0.61 \pm 0.01$ \\
  & Geo2Vec & $0.62 \pm 0.00$ & $0.63 \pm 0.00$ & $0.60 \pm 0.00$ & $0.61 \pm 0.00$ & $\mathbf{0.62} \pm 0.01$ \\
\midrule
\multirow{4}{*}{Munich} 
  & JEPA & $0.58 \pm 0.01$ & $0.65 \pm 0.00$ & $0.61 \pm 0.00$ & $0.57 \pm 0.00$ & $0.59 \pm 0.02$ \\
  & MAE & $0.56 \pm 0.02$ & $0.57 \pm 0.01$ & $0.55 \pm 0.00$ & $0.51 \pm 0.00$ & $\mathbf{0.64} \pm 0.06$ \\
  & Poly2Vec & $\mathbf{0.62} \pm 0.00$ & $\mathbf{0.73} \pm 0.00$ & $\mathbf{0.67} \pm 0.00$ & $\mathbf{0.64} \pm 0.00$ & $0.61 \pm 0.01$ \\
  & Geo2Vec & $0.58 \pm 0.01$ & $0.64 \pm 0.00$ & $0.61 \pm 0.00$ & $0.57 \pm 0.00$ & $0.60 \pm 0.02$ \\
\midrule
\multirow{4}{*}{Singapore} 
  & JEPA & $0.47 \pm 0.01$ & $0.59 \pm 0.00$ & $0.61 \pm 0.00$ & $0.52 \pm 0.01$ & $0.42 \pm 0.02$ \\
  & MAE & $0.23 \pm 0.32$ & $0.63 \pm 0.00$ & $0.52 \pm 0.10$ & $0.16 \pm 0.22$ & $0.40 \pm 0.54$ \\
  & Poly2Vec & $0.47 \pm 0.01$ & $\mathbf{0.66} \pm 0.00$ & $\mathbf{0.62} \pm 0.01$ & $\mathbf{0.54} \pm 0.02$ & $0.42 \pm 0.03$ \\
  & Geo2Vec & $\mathbf{0.50} \pm 0.02$ & $0.63 \pm 0.00$ & $0.61 \pm 0.00$ & $0.52 \pm 0.00$ & $\mathbf{0.49} \pm 0.03$ \\
\bottomrule
\end{tabular}%
}
\end{table*}

First, unlike the polygon evaluation where a single approach (i.e., Geo2Vec) outperformed universally, the quality assessment of street network reveals a dynamic performance behavior between Poly2Vec and Geo2Vec. For both overshoot and undershoot errors, Poly2Vec generally emerges as a preferred approach in Los Angeles and Munich. For instance, in detecting overshoots in Los Angeles, Poly2Vec achieves the highest F1 of 0.64 and an AUC of 0.70. Similarly, for undershoot detection in Munich, Poly2Vec secures a higher F1 of 0.61 and an accuracy of 0.62. However, Geo2Vec proves to be the most robust approach for the complex Singapore dataset, achieving the best F1 for both overshoots (i.e., 0.50) and undershoots (i.e., 0.51), successfully outperforming three other baselines where they struggled.

Second, this quantitative results emphasis that detecting topological errors in street networks is an inherently more challenging task than identifying overlapping building polygons. Though the polygon classification results frequently achieved F1 well above 0.90, the performance metrics for street networks are significantly lower across all three study areas. The upperbound F1 across both overshoot and undershoot is about 0.64, and the highest accuracy caps at 0.67. This significant performance gap suggests that polyline-based vector data, characterized by one-dimensional coordinate sequences and precise junction connections, consist of highly complex spatial representations that are much harder for existing SRL approaches to fully address.

Third, the MAE baseline struggles significantly with learning effective neural representations for street networks, resulting in both poor predictive performance and severe training instability. This fact is more evident in the Singapore dataset, where the MAE approach completely fails to learn effective representations for overshoots and undershoots. For overshoot detection in Singapore, MAE gives very low averag F1 of 0.23 coupled with a massive standard deviation of 0.32. For undershoot detection, it is even worse, dropping to an F1 of 0.12 with a standard deviation of 0.26. In contrast, the FFT-based and SDF-based approaches like JEPA, Poly2Vec, and Geo2Vec achieves stable classification results, with standard deviations generally remaining between 0.00 and 0.02 across almost all metrics and cities.

\begin{table*}[!htbp]
\centering
\small
\caption{Performances of \textbf{Undershoot Street Network} Detection in three study areas (with best result in bold, and all metrics in Mean $\pm$ Std, from 5 runs)}
\label{tab:undershoot_contrastive}
\resizebox{\textwidth}{!}{%
\begin{tabular}{@{} l l *{5}{c} @{}}
\toprule
City & Base & F1 & AUC & Accuracy & Precision & Recall \\
\midrule
\multirow{4}{*}{Los Angeles} 
  & JEPA & $0.61 \pm 0.00$ & $0.60 \pm 0.00$ & $0.57 \pm 0.00$ & $0.59 \pm 0.00$ & $\mathbf{0.63} \pm 0.01$ \\
  & MAE & $0.59 \pm 0.05$ & $0.60 \pm 0.00$ & $0.57 \pm 0.02$ & $0.60 \pm 0.01$ & $0.58 \pm 0.10$ \\
  & Poly2Vec & $\mathbf{0.63} \pm 0.01$ & $\mathbf{0.65} \pm 0.00$ & $\mathbf{0.60} \pm 0.00$ & $\mathbf{0.63} \pm 0.01$ & $\mathbf{0.63} \pm 0.02$ \\
  & Geo2Vec & $0.62 \pm 0.01$ & $\mathbf{0.65} \pm 0.00$ & $\mathbf{0.60} \pm 0.00$ & $0.62 \pm 0.00$ & $0.62 \pm 0.02$ \\
\midrule
\multirow{4}{*}{Munich} 
  & JEPA & $0.59 \pm 0.00$ & $0.63 \pm 0.00$ & $0.59 \pm 0.00$ & $0.55 \pm 0.00$ & $\mathbf{0.63} \pm 0.01$ \\ 
  & MAE & $0.56 \pm 0.02$ & $0.57 \pm 0.01$ & $0.55 \pm 0.01$ & $0.51 \pm 0.01$ & $0.64 \pm 0.06$ \\
  & Poly2Vec & $\mathbf{0.61} \pm 0.01$ & $\mathbf{0.67} \pm 0.00$ & $\mathbf{0.62} \pm 0.00$ & $\mathbf{0.59} \pm 0.01$ & $\mathbf{0.63} \pm 0.04$ \\
  & Geo2Vec & $0.59 \pm 0.00$ & $0.66 \pm 0.00$ & $0.61 \pm 0.00$ & $0.58 \pm 0.00$ & $0.61 \pm 0.01$ \\
\midrule
\multirow{4}{*}{Singapore} 
  & JEPA & $0.42 \pm 0.05$ & $0.56 \pm 0.01$ & $0.54 \pm 0.01$ & $0.43 \pm 0.02$ & $0.41 \pm 0.08$ \\
  & MAE & $0.12 \pm 0.26$ & $0.56 \pm 0.01$ & $0.56 \pm 0.08$ & $0.08 \pm 0.18$ & $0.20 \pm 0.45$ \\
  & Poly2Vec & $0.42 \pm 0.06$ & $0.59 \pm 0.01$ & $0.58 \pm 0.01$ & $0.48 \pm 0.02$ & $0.38 \pm 0.09$ \\
  & Geo2Vec & $\mathbf{0.51} \pm 0.01$ & $\mathbf{0.64} \pm 0.00$ & $\mathbf{0.59} \pm 0.01$ & $\mathbf{0.51} \pm 0.01$ & $\mathbf{0.51} \pm 0.02$ \\
\bottomrule
\end{tabular}%
}
\end{table*}

\begin{figure}[!htbp]
    \centering
    \includegraphics[width=\linewidth]{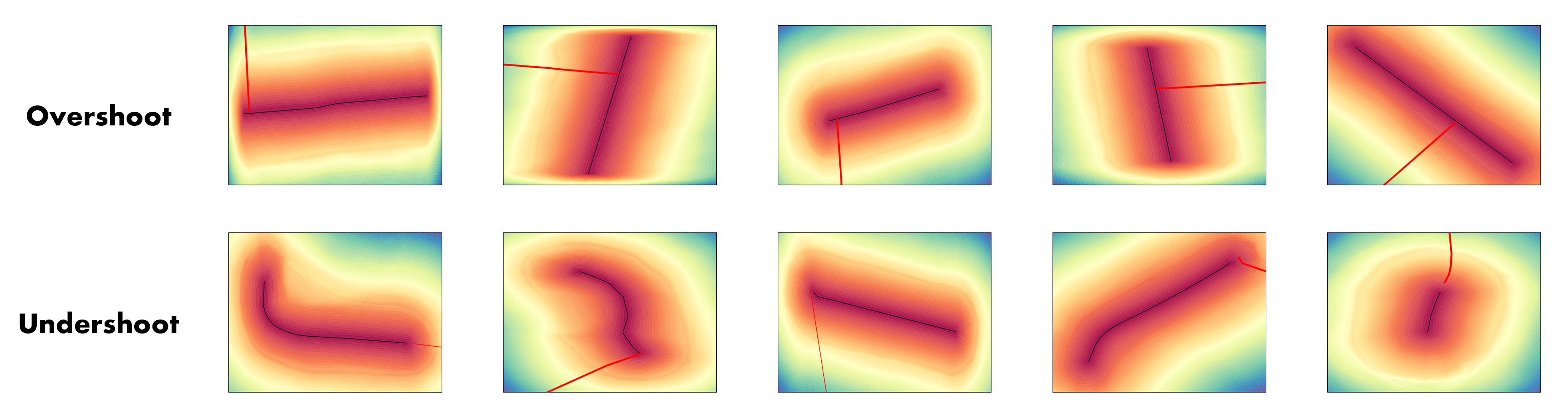}
    \caption{Selected visualization of learned SDF representing overshoot and undershoot errors in street networks.}
    \label{fig:polyline_examples}
\end{figure}

Visualization of the overshoot and undershoot SDF maps in Figure \ref{fig:polyline_examples} provides additional information to understand and explain why the topological errors of street networks is considered more difficult than overlapping polygons. One can clearly notice that the trivial difference between overshoot and undershoot polylines as well as their SDF representations. Despite the difficult nature of this task, we argued that existing SRL approaches lack of an internal mechanism to represent connectivity and direction, which can be a nice future direction in GeoAI and SRL to enhance its capability in detecting polyline topological and connectivity errors.

\subsection{Ablation Study}

To further validate Topo4Vec's design and limitation, we designed three ablation studies. First, we compared contrastive learning approaches against a naive softmax baseline to examine the actual performance gains from the contrastive learning objective. Second, we examined cross-city generalizability of Topo4Vec by evaluating how well the learned neural representations transfer across distinct urban morphologies without of additional training. 

\subsubsection{Contrastive and Naive Classifiers}

\begin{figure}[!t]
    \centering
    \includegraphics[width=\linewidth]{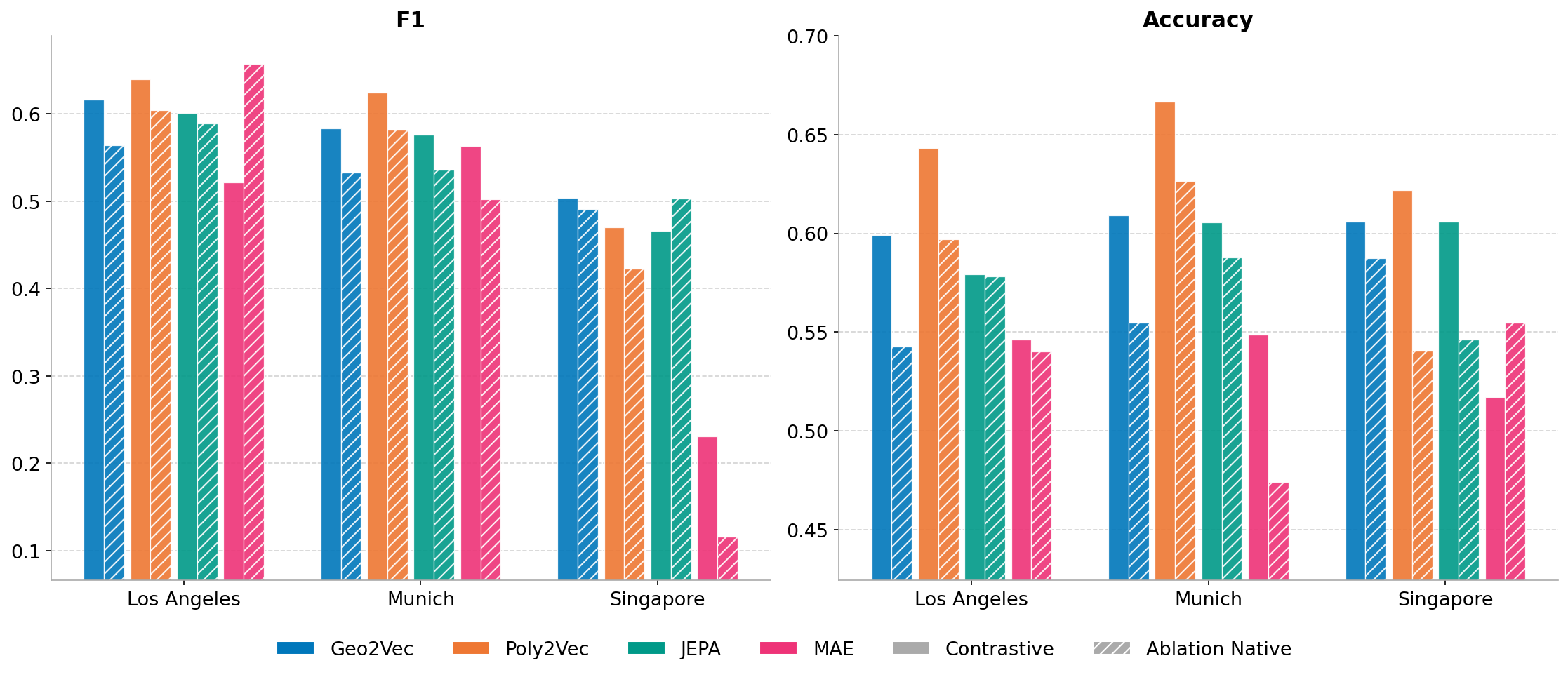}
    \caption{Ablation studies comparing contrastive approaches against naive classifiers for overshoot error detection across the three study areas.}
    \label{fig:ablation_overshoot}
\end{figure}

\begin{figure}[!t]
    \centering
    \includegraphics[width=\linewidth]{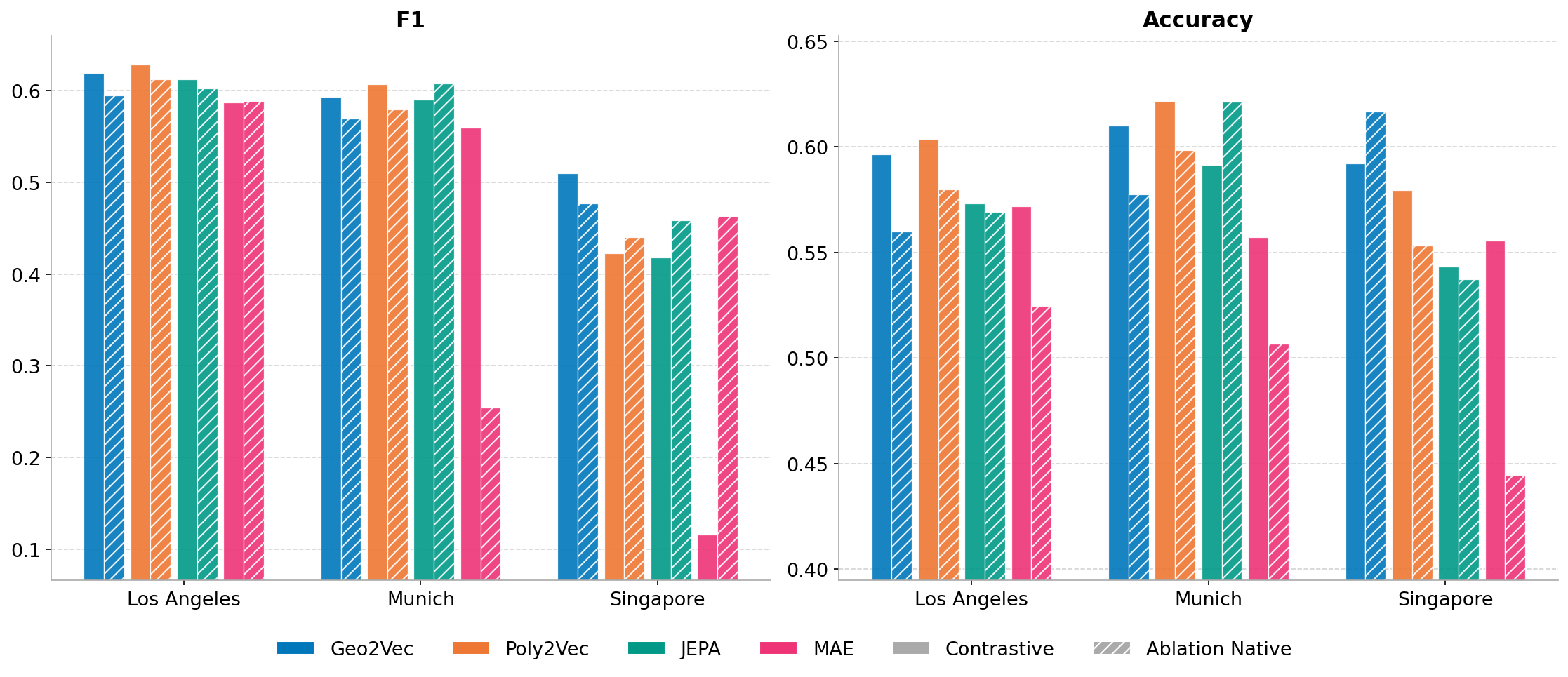}
    \caption{Ablation studies comparing contrastive approaches against naive classifiers for undershoot error detection across the three study areas.}
    \label{fig:ablation_undershoot}
\end{figure}

To assess the impact of the contrastive learning objectives, we designed an ablation study, especially for the more challenging task of street network topological errors (e.g., Overshoot and Undershoot).  Figure \ref{fig:ablation_overshoot} and \ref{fig:ablation_undershoot} compare the contrastive approaches (i.e., solid bars) with their naive classifier ablations (i.e., hatched bars). The contrastive approach shows a clear advantage in both F1 and Accuracy, particularly for Geo2Vec and Poly2Vec. This observation confirms that optimizing using contrastive topological representations contribute to handle complex spatial structures much better than standard softmax classification.

Furthermore, this ablation shows that a contrastive objective cannot fully fix poor spatial representation capabilities. This is obvious in the MAE baseline's performance on the complex Singapore dataset. For both error types (ie., undershoots and overshoots), the contrastive MAE model struggles with low F1 plunging near 0.1 and 0.2. Therefore, adding a contrastive objective onto less capable SRL approach simply lead to worse results, saying that contrastive learning is best combined with a robust spatial encoder like Geo2Vec to be really effective.

\subsubsection{Geographical Generalizability Test}

To test the geographical generalizability of Topo4Vec, we conducted a second ablation study by applying those Topo4Vec models trained on Los Angeles (LA) and Munich (MUN) directly to Singapore (SG) considering its more challenging urban morphologies. 

Table \ref{tab:transfer_results_rounded} elaborates on the cross-city transfer performance for both overshoot and undershoot, respectively. For overshoots, poly2vec slightly outperforms Geo2Vec, maintaining steady F1 of 0.55 (LA to SG) and 0.56 (MUN to SG). For undershoots, the trend differs, and Geo2Vec proves to be more adaptable. As a evidence, Geo2Vec outperforms poly2vec when transferring from LA to SG (i.e., achieving an F1 of 0.53 over 0.48) and give the highest accuracy of 0.64 when transferring from Munich. 

Nevertheless, one can observe that overall performance drops significantly compared to the models' baseline intra-city results. This fact highlights a fundamental reality that street network topologies and their corresponding error patterns are deeply tied to local urban design, making geographical generalizability an inherently challenging and desiring feature for any SRL approaches \citep{li2023rethink}.

\begin{table*}[!t]
\centering
\caption{Performance of Cross-city Transfer from Source Cities (i.e., Los Angeles, Munich) to Target City (i.e., Singapore) for Overshoot and Undershoot (all metrics in Mean $\pm$ Std, from 5 runs). }
\label{tab:transfer_results_rounded}
\resizebox{\textwidth}{!}{
\begin{tabular}{lllccccc}
\toprule
\textbf{Setting} & \textbf{Model} & \textbf{Transfer} & \textbf{Accuracy} & \textbf{Precision} & \textbf{Recall} & \textbf{F1} & \textbf{AUC} \\
\midrule
\multirow{4}{*}{\textbf{Overshoot}} 
& poly2vec & LA $\rightarrow$ SG  & $0.64 \pm 0.01$ & $0.52 \pm 0.01$ & $0.59 \pm 0.02$ & $0.55 \pm 0.01$ & $0.66 \pm 0.00$ \\
& poly2vec & MUN $\rightarrow$ SG & $0.60 \pm 0.00$ & $0.52 \pm 0.00$ & $0.61 \pm 0.01$ & $0.56 \pm 0.01$ & $0.63 \pm 0.00$ \\
& Geo2Vec  & LA $\rightarrow$ SG  & $0.56 \pm 0.00$ & $0.43 \pm 0.00$ & $0.61 \pm 0.01$ & $0.51 \pm 0.01$ & $0.59 \pm 0.00$ \\
& Geo2Vec  & MUN $\rightarrow$ SG & $0.55 \pm 0.01$ & $0.46 \pm 0.02$ & $0.30 \pm 0.01$ & $0.36 \pm 0.01$ & $0.54 \pm 0.00$ \\
\midrule
\multirow{4}{*}{\textbf{Undershoot}} 
& poly2vec & LA $\rightarrow$ SG  & $0.54 \pm 0.01$ & $0.44 \pm 0.01$ & $0.53 \pm 0.02$ & $0.48 \pm 0.01$ & $0.58 \pm 0.00$ \\
& poly2vec & MUN $\rightarrow$ SG & $0.54 \pm 0.01$ & $0.45 \pm 0.01$ & $0.55 \pm 0.01$ & $0.50 \pm 0.01$ & $0.59 \pm 0.00$ \\
& Geo2Vec  & LA $\rightarrow$ SG  & $0.56 \pm 0.01$ & $0.46 \pm 0.01$ & $0.62 \pm 0.02$ & $0.53 \pm 0.01$ & $0.58 \pm 0.00$ \\
& Geo2Vec  & MUN $\rightarrow$ SG & $0.64 \pm 0.00$ & $0.57 \pm 0.00$ & $0.43 \pm 0.01$ & $0.49 \pm 0.01$ & $0.65 \pm 0.00$ \\
\bottomrule
\end{tabular}
}
\end{table*}

\subsubsection{Train and Test Performance Gap}

\begin{figure}[H]
    \centering
    \includegraphics[width=\linewidth]{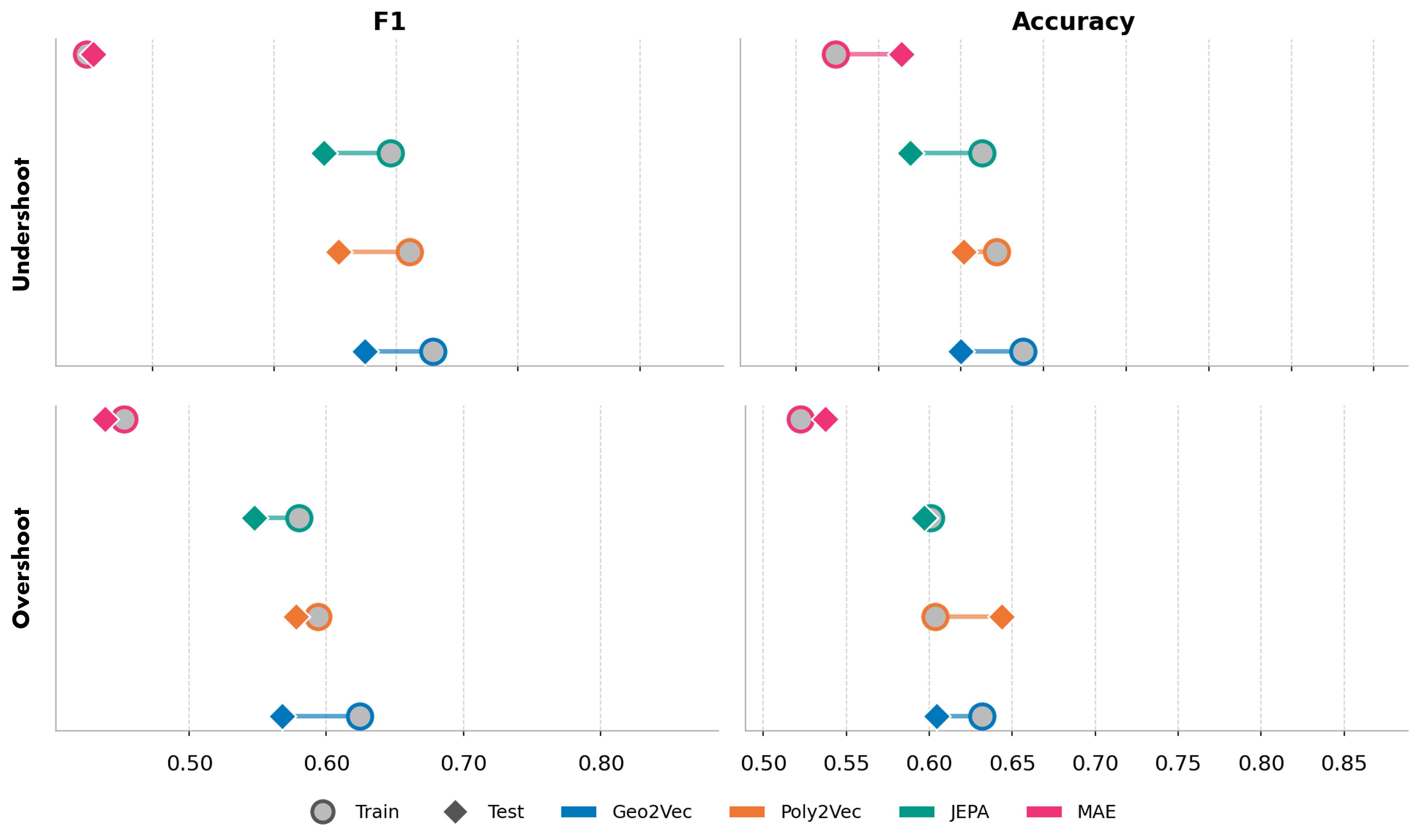}
    \caption{Train and Test Performance Gap visualized using dumbbell plot to illustrate the generalization robustness of different models for both overshoot and undershoot detection.}
    \label{fig:traintest_gap}
\end{figure}

The last ablation study investigate the performance gap between train and test for overshoot and undershoot in street networks. Herein, as the goal to understand the true difficulty, we analysed this gap by averaging model performances over three cities. Figure \ref{fig:traintest_gap} visualizes the generalization robustness of the different Topo4Vec models, where the length of each line segment refers to the magnitude of the performance degradation from training to testing. The  dumbbell plot in Figure \ref{fig:traintest_gap} reveals the varying degree of generalization robustness across the different models and topological error types. 

In the one hand, models exhibiting extended line segments demonstrate a bigger divergence between training and testing performance. This clear performance gap indicates a strong tendency to memorize specific local geometric pattern and coordinate sequences from the training data rather than learning broadly generalizable spatial features.

In the other hand, shorter line segments reflect better generalization capabilities. Models that maintain a tight alignment between their training and testing performance prove that they are not simply relying on memorization. Instead, they effectively capture the underlying, generalizable topological invariants of the street networks, ensuring that their testing results remains stable and reliable when applied to unseen urban environments. Therefore, an ideal scenario will need to balance the overall performance (e.g., F1 and Accuracy) and the robustness (e.g., train and test gap).


\section{Discussion}

This study introduced Topo4Vec, an automated GeoAI framework, designed for topological quality assessment of vector data (e.g., such as polygons and polylines) using advanced SRL approaches. By evaluating the framework across three diverse urban environments (e.g., Singapore, Los Angeles, and Munich), we seek to address three core RQs and identify potential future research directions.

\subsection{RQ1: Synergizing Neural Representation and Contrastive Learning}
Our findings demonstrate that Topo4Vec combines SRL approaches (i.e., Geo2Vec) with contrastive learning to effectively and automatically detect fine-scale topological errors. Moreover, the SDF feature of Geo2Vec provides a promising solution to embed distance and shape morphology into the latent space, while the pairwise contrastive objective explicitly targets hard negatives to prevent the memorization of local coordinate sequences. The ablation studies confirm this approach acts as a powerful learning tools that contribute to close the train-test performance gap, especially in the challenging detection of street network overshoot and undershoot.

Furthermore, by encoding complex non-Euclidean geometries and topological relations into a continuous, differentiable latent space, Topo4Vec offers a profound opportunity to revisit the central theories of vector data analysis in GIS. Traditional GIS relies heavily on discrete, rigid geometric predicates (e.g., DE-9IM model \citep{egenhofer1993critical}) which struggles with massive computational overhead and contextual ambiguity at a global scale. Transitioning to the learning-based, neural representations offers a stimulating opportunity for more scalable and autonomous GIS analysis, enabling GeoAI systems to inherently learn complex and potentially unknown spatial relationships without exhaustive, human-coded rules.

\subsection{RQ2: Topological Error Simulation and Quality Assessment}
In Topo4Vec, we designed an automated topological error simulation workflow, which is considered a highly flexible alternative than classic labor-intensive manual annotations, especially with the massive amount of vector data. By randomly generating realistic and balanced geometric anomalies, such as stochastic overlapping polygons and constrained network over- and undershoot, the simulation workflow can significantly reduce the burden of manual annotations. Moreover, the strong intra-city performance, highlighted by a peak F1 of 0.99 and AUC of 1.00 for Munich polygon overlaps, empirically validates that models trained on simulated data can accurately assess real-world topological data quality as well. 

This automated, flexible simulation workflow holds immense potential for real-world practice, particularly within fast-growing geospatial data ecosystems such as Grab Maps \footnote{https://grabmaps.grab.com/}, Uber Maps, and those location-based services (LBS) platforms. These platforms depend heavily on massive, continuously updating crowdsourcing street networks and building footprints where manual quality assurance is inherently inefficient and critically slow. By leveraging the Topo4Vec framework, these commercial platforms can deploy continuous, autonomous vector data quality assessment practices to detect street network undershoots, overshoots, or overlapping building footprints thereby ensuring safe, efficient, and reliable navigation and routing at scale.

\subsection{RQ3: Geographical Generalizability and Urban Morphological Context}
Last but not least, our experiment results reveal that model performance and generalizability are heavily influenced by local urban morphological context. Although the Topo4Vec framework proved to be robust in the rectilinear grids of Los Angeles and the classic layouts of Munich, the complex urban landscape of Singapore's high-density housing post a significant challenge. Furthermore, cross-city transfer tests showed significant performance drops, specifically for overshoot and undershoot of street networks, highlighting that urban morphologies which are deeply inherent in the vector data remain a critical barrier for geographical generalizability. 

Addressing this generalizability problem is not only a research question, it is also a vital prerequisite for deploying scalable and reproducible deployment of GeoAI models in a global scale \citep{li2024geoai}. For instance, as ride-hailing and commercial logistics companies expand across borders, their routing systems must be capable of rapidly adapting to entirely new urban layouts without the need for extensive retraining. Therefore, advancing generalizable SRL will be critical in facilitating the vision of autonomous GIS, for example to ensure that GeoAI models can seamlessly transfer foundational geographic knowledge learned from grid-like American cities to the dense, unique urban landscape of Southeast Asian cities.

\section{Conclusion}
In this paper, we present TopoVec, a novel and automated GeoAI framework, that leverages spatial representation learning (SRL) to assess topological quality of vector data, such as overlapping polygons, overshoot, and undershoot street networks. By reducing the need for manual labeling process through topological error simulation, TopoVec offers a promising solution for automated, scalable, and GeoAI-based vector data quality assurance. We evaluate the proposed framework across three diverse urban morphologies of Singapore, Los Angeles, and Munich. Preliminary experiment results comparing with state-of-the-art SRL approaches demonstrate that TopoVec is highly effective and robust at identifying overlapping building polygons, achieving a promising accuracy of 0.98 and F1 scores of 0.99. However, the evaluation also points out that detecting topological errors in street networks, specifically overshoots and undershoots, remains a challenging task, as current SRL approaches struggle to inherently consider complex spatial connectivity and network direction in the neural representation, which leaves future works a lot of potential and room to improve.

Despite the challenge, the potential of SRL to encode complex, native vector geometries into standard and continuous latent neural representation presents a stimulating opportunity for accurate, scalable, and automatic vector data quality assessment, which will greatly support global spatial data infrastructure ecosystems. Last but not least, as future work still have to overcome the limitations of cross-scale topological geometry encoding and cross-city generalizability, advanced SRL approach show a great potential to fundamentally revisit classic, rule-based GIS vector data analysis aligning with the overall research trend of GeoAI in the GIScience field.

\section*{Acknowledgments}

This research was supported by the project ``Geospatial Artificial Intelligence for Climate Resilient Urban Environment”, funded by the National University of Singapore under the Start Up Grant (E-109-00-0036-01). This works is also part of project ``AI4OpenGeodataQuality: A Toolkit and Framework for Continuous Large-Scale Quality Measurement of Open Geodata with Efficient AI and Multimodal Remote Sensing" funded by the Deutsche Forschungs Gemeinschaft (DFG) (Grant Nos.563303373).

\section*{Generative AI Use Disclosure Statement}
During the preparation of this work, the authors used Gemini 3.5 to improve the readability and language of the manuscript. After using this tool, the authors reviewed and edited the content as needed and take full responsibility for the content of the published article.

\section*{Data and Codes Availability Statement}

The code and data used in the paper are made openly available in \mbox{\url{https://figshare.com/s/612148eeb4bccadbd715}}.

\section*{Disclosure statement}
The authors report there are no competing interests to declare.

\bibliographystyle{tfv}
\bibliography{interacttfvsample}

@article{goodchild2007citizens,
  author  = {Goodchild, M. F.},
  title   = {Citizens as sensors: the world of volunteered geography},
  journal = {GeoJournal},
  year    = {2007},
  volume  = {69},
  number  = {4},
  pages   = {211--221}
}

@article{senaratne2017review,
  author  = {Senaratne, H. and Mobasheri, A. and Ali, A. L. and Capineri, C. and Haklay, M.},
  title   = {A review of volunteered geographic information quality assessment methods},
  journal = {International Journal of Geographical Information Science},
  year    = {2017},
  volume  = {31},
  number  = {1},
  pages   = {139--167}
}

@inproceedings{maas2013rectifier,
  title={Rectifier nonlinearities improve neural network acoustic models},
  author={Maas, Andrew L and Hannun, Awni Y and Ng, Andrew Y and others},
  booktitle={Proc. icml},
  volume={30},
  number={1},
  pages={3},
  year={2013},
  organization={Atlanta, GA}
}

@article{haklay2010how,
  author  = {Haklay, M.},
  title   = {How good is volunteered geographical information? A comparative study of OpenStreetMap and Ordnance Survey datasets},
  journal = {Environment and planning B: Planning and design},
  year    = {2010},
  volume  = {37},
  number  = {4},
  pages   = {682--703}
}

@article{barron2014comprehensive,
  author  = {Barron, C. and Neis, P. and Zipf, A.},
  title   = {A comprehensive framework for intrinsic OpenStreetMap quality analysis},
  journal = {Transactions in GIS},
  year    = {2014},
  volume  = {18},
  number  = {6},
  pages   = {877--895}
}

@article{egenhofer1991point,
  author  = {Egenhofer, M. J. and Franzosa, R. D.},
  title   = {Point-set topological spatial relations},
  journal = {International Journal of Geographical Information Systems},
  year    = {1991},
  volume  = {5},
  number  = {2},
  pages   = {161--174}
}

@incollection{clementini1993small,
  author    = {Clementini, E. and Di Felice, P. and Van Oosterom, P.},
  title     = {A small set of formal topological relationships suitable for end-user interaction},
  booktitle = {Advances in spatial databases},
  year      = {1993},
  pages     = {277--295}
}

@article{fan2014quality,
  author  = {Fan, H. and Zipf, A. and Fu, Q. and Neis, P.},
  title   = {Quality assessment for building footprints data on OpenStreetMap},
  journal = {International Journal of Geographical Information Science},
  year    = {2014},
  volume  = {28},
  number  = {4},
  pages   = {700--719}
}

@article{girres2010quality,
  author  = {Girres, J. F. and Touya, G.},
  title   = {Quality assessment of the French OpenStreetMap dataset},
  journal = {Transactions in GIS},
  year    = {2010},
  volume  = {14},
  number  = {4},
  pages   = {435--459}
}

@article{wu2020comprehensive,
  author  = {Wu, Z. and Pan, S. and Chen, F. and Long, G. and Zhang, C. and Philip, S. Y.},
  title   = {A comprehensive survey on graph neural networks},
  journal = {IEEE Transactions on Neural Networks and Learning Systems},
  year    = {2020},
  volume  = {32},
  number  = {1},
  pages   = {4--24}
}

@inproceedings{geo2vec_paper,
  title={Geo2vec: Shape-and distance-aware neural representation of geospatial entities},
  author={Chu, Chen and Shahabi, Cyrus},
  booktitle={Proceedings of the AAAI Conference on Artificial Intelligence},
  volume={40},
  number={23},
  pages={18985--18993},
  year={2026}
}

@article{tuia2016domain,
  author  = {Tuia, D. and Persello, C. and Bruzzone, L.},
  title   = {Domain adaptation for the classification of remote sensing data: An overview of recent advances},
  journal = {IEEE Geoscience and Remote Sensing Magazine},
  year    = {2016},
  volume  = {4},
  number  = {2},
  pages   = {41--57}
}

@inproceedings{li2023rethink,
  title={Rethink geographical generalizability with unsupervised self-attention model ensemble: A case study of openstreetmap missing building detection in africa},
  author={Li, Hao and Wang, Jiapan and Zollner, Johann Maximilian and Mai, Gengchen and Lao, Ni and Werner, Martin},
  booktitle={Proceedings of the 31st ACM International Conference on Advances in Geographic Information Systems},
  pages={1--9},
  year={2023}
}

@article{goodchild2021replication,
  title={Replication across space and time must be weak in the social and environmental sciences},
  author={Goodchild, Michael F and Li, Wenwen},
  journal={Proceedings of the National Academy of Sciences},
  volume={118},
  number={35},
  pages={e2015759118},
  year={2021},
  publisher={National Acad Sciences}
}

@inproceedings{park2019deepsdf,
  title={Deepsdf: Learning continuous signed distance functions for shape representation},
  author={Park, Jeong Joon and Florence, Peter and Straub, Julian and Newcombe, Richard and Lovegrove, Steven},
  booktitle={Proceedings of the IEEE/CVF conference on computer vision and pattern recognition},
  pages={165--174},
  year={2019}
}

@article{biljecki2022global,
  title={Global building morphology indicators},
  author={Biljecki, Filip and Chow, Yoong Shin},
  journal={Computers, Environment and Urban Systems},
  volume={95},
  pages={101809},
  year={2022},
  publisher={Elsevier}
}

@article{liu2022review,
  title={A review of spatially-explicit GeoAI applications in Urban Geography},
  author={Liu, Pengyuan and Biljecki, Filip},
  journal={International Journal of Applied Earth Observation and Geoinformation},
  volume={112},
  pages={102936},
  year={2022},
  publisher={Elsevier}
}

@article{hong2023cross,
  title={Cross-city matters: A multimodal remote sensing benchmark dataset for cross-city semantic segmentation using high-resolution domain adaptation networks},
  author={Hong, Danfeng and Zhang, Bing and Li, Hao and Li, Yuxuan and Yao, Jing and Li, Chenyu and Werner, Martin and Chanussot, Jocelyn and Zipf, Alexander and Zhu, Xiao Xiang},
  journal={Remote Sensing of Environment},
  volume={299},
  pages={113856},
  year={2023},
  publisher={Elsevier}
}

@article{xu2023universal,
  title={Universal domain adaptation for remote sensing image scene classification},
  author={Xu, Qingsong and Shi, Yilei and Yuan, Xin and Zhu, Xiao Xiang},
  journal={IEEE Transactions on Geoscience and Remote Sensing},
  volume={61},
  pages={1--15},
  year={2023},
  publisher={IEEE}
}

@article{liu2025graph,
  title={A graph neural network for small-area estimation: integrating spatial regularisation, heterogeneous spatial units, and Bayesian inference},
  author={Liu, Pengyuan and Chen, Yang and Liang, Xiucheng and Li, Hao and Biljecki, Filip and Stouffs, Rudi},
  journal={International Journal of Geographical Information Science},
  pages={1--39},
  year={2025},
  publisher={Taylor \& Francis}
}

@article{herfort2023spatio,
  title={A spatio-temporal analysis investigating completeness and inequalities of global urban building data in OpenStreetMap},
  author={Herfort, Benjamin and Lautenbach, Sven and Porto de Albuquerque, Jo{\~a}o and Anderson, Jennings and Zipf, Alexander},
  journal={Nature Communications},
  volume={14},
  number={1},
  pages={3985},
  year={2023},
  publisher={Nature Publishing Group UK London}
}

@article{li2022improving,
  title={Improving OpenStreetMap missing building detection using few-shot transfer learning in sub-Saharan Africa},
  author={Li, Hao and Herfort, Benjamin and Lautenbach, Sven and Chen, Jiaoyan and Zipf, Alexander},
  journal={Transactions in GIS},
  volume={26},
  number={8},
  pages={3125--3146},
  year={2022},
  publisher={Wiley Online Library}
}

@article{li2020exploration,
  title={Exploration of OpenStreetMap missing built-up areas using twitter hierarchical clustering and deep learning in Mozambique},
  author={Li, Hao and Herfort, Benjamin and Huang, Wei and Zia, Mohammed and Zipf, Alexander},
  journal={ISPRS Journal of Photogrammetry and Remote Sensing},
  volume={166},
  pages={41--51},
  year={2020},
  publisher={Elsevier}
}

@article{li2025cross,
  title={Cross-view geolocalization and disaster mapping with street-view and VHR satellite imagery: A case study of Hurricane IAN},
  author={Li, Hao and Deuser, Fabian and Yin, Wenping and Luo, Xuanshu and Walther, Paul and Mai, Gengchen and Huang, Wei and Werner, Martin},
  journal={ISPRS Journal of Photogrammetry and Remote Sensing},
  volume={220},
  pages={841--854},
  year={2025},
  publisher={Elsevier}
}

@inproceedings{egenhofer1993critical,
  title={A critical comparison of the 4-intersection and 9-intersection models for spatial relations: formal analysis},
  author={Egenhofer, Max J and Sharma, Jayant and Mark, David M and others},
  booktitle={AUTOCARTO-CONFERENCE-},
  pages={1--1},
  year={1993},
  organization={ASPRS AMERICAN SOCIETY FOR PHOTOGRAMMETRY AND}
}

@article{li2023autonomous,
  title={Autonomous GIS: the next-generation AI-powered GIS},
  author={Li, Zhenlong and Ning, Huan},
  journal={International Journal of Digital Earth},
  volume={16},
  number={2},
  pages={4668--4686},
  year={2023},
  publisher={Taylor \& Francis}
}

@article{zhang2022deep,
  title={Deep learning for processing and analysis of remote sensing big data: A technical review},
  author={Zhang, Xin and Zhou, Ya’nan and Luo, Jiancheng},
  journal={Big Earth Data},
  volume={6},
  number={4},
  pages={527--560},
  year={2022},
  publisher={Taylor \& Francis}
}

@inproceedings{mai2024srl,
  title={SRL: Towards a general-purpose framework for spatial representation learning},
  author={Mai, Gengchen and Yao, Xiaobai and Xie, Yiqun and Rao, Jinmeng and Li, Hao and Zhu, Qing and Li, Ziyuan and Lao, Ni},
  booktitle={Proceedings of the 32nd ACM international conference on advances in geographic information systems},
  pages={465--468},
  year={2024}
}

@inproceedings{wang2024random,
  title={Random Affine Transformation Feature Representation Learning for Fast Polygon Retrieval},
  author={Wang, Zhangyu and Du, Xingyi and Li, Hao and Werner, Martin},
  booktitle={Proceedings of the 3rd ACM SIGSPATIAL International Workshop on Searching and Mining Large Collections of Geospatial Data},
  pages={22--28},
  year={2024}
}

@article{siampou2025poly2vec,
  title={Poly2vec: Polymorphic fourier-based encoding of geospatial objects for geoai applications},
  author={Siampou, Maria Despoina and Li, Jialiang and Krumm, John and Shahabi, Cyrus and Lu, Hua},
  journal={Proceedings of machine learning research},
  volume={267},
  pages={55511},
  year={2025}
}

@article{huang2022estimating,
  title={Estimating urban functional distributions with semantics preserved POI embedding},
  author={Huang, Weiming and Cui, Lizhen and Chen, Meng and Zhang, Daokun and Yao, Yao},
  journal={International Journal of Geographical Information Science},
  volume={36},
  number={10},
  pages={1905--1930},
  year={2022},
  publisher={Taylor \& Francis}
}

@article{liu2020dynamic,
  title={Dynamic spatial-temporal representation learning for traffic flow prediction},
  author={Liu, Lingbo and Zhen, Jiajie and Li, Guanbin and Zhan, Geng and He, Zhaocheng and Du, Bowen and Lin, Liang},
  journal={IEEE Transactions on Intelligent Transportation Systems},
  volume={22},
  number={11},
  pages={7169--7183},
  year={2020},
  publisher={IEEE}
}

@article{hu2023recognizing,
  title={Recognizing mixed urban functions from human activities using representation learning methods},
  author={Hu, Junjie and Gao, Yong and Wang, Xuechen and Liu, Yu},
  journal={International Journal of Digital Earth},
  volume={16},
  number={1},
  pages={289--307},
  year={2023},
  publisher={Taylor \& Francis}
}

@article{liu2025representation,
  title={Representation learning for geospatial data},
  author={Liu, Yu and Wang, Xuechen and Wang, Yidan and Huang, Fei and Huang, Yingjing and Li, Yong and Zhang, Weiyu and Gong, Shuhui and Mai, Gengchen and Yao, Yao and others},
  journal={Annals of GIS},
  volume={31},
  number={4},
  pages={557--583},
  year={2025},
  publisher={Taylor \& Francis}
}

@inproceedings{chu2019geo,
  title={Geo-aware networks for fine-grained recognition},
  author={Chu, Grace and Potetz, Brian and Wang, Weijun and Howard, Andrew and Song, Yang and Brucher, Fernando and Leung, Thomas and Adam, Hartwig},
  booktitle={Proceedings of the IEEE/CVF International Conference on Computer Vision Workshops},
  pages={0--0},
  year={2019}
}

@article{mai2022review,
  title={A review of location encoding for GeoAI: methods and applications},
  author={Mai, Gengchen and Janowicz, Krzysztof and Hu, Yingjie and Gao, Song and Yan, Bo and Zhu, Rui and Cai, Ling and Lao, Ni},
  journal={International Journal of Geographical Information Science},
  volume={36},
  number={4},
  pages={639--673},
  year={2022},
  publisher={Taylor \& Francis}
}

@article{hu2024five,
  title={A five-year milestone: reflections on advances and limitations in GeoAI research},
  author={Hu, Yingjie and Goodchild, Michael and Zhu, A-Xing and Yuan, May and Aydin, Orhun and Bhaduri, Budhendra and Gao, Song and Li, Wenwen and Lunga, Dalton and Newsam, Shawn},
  journal={Annals of GIS},
  volume={30},
  number={1},
  pages={1--14},
  year={2024},
  publisher={Taylor \& Francis}
}

@article{goodchild2013quality,
  title={The quality of big (geo) data},
  author={Goodchild, Michael F},
  journal={Dialogues in Human Geography},
  volume={3},
  number={3},
  pages={280--284},
  year={2013},
  publisher={SAGE Publications Sage UK: London, England}
}

@article{li2022geoai,
  title={GeoAI for large-scale image analysis and machine vision: Recent progress of artificial intelligence in geography},
  author={Li, Wenwen and Hsu, Chia-Yu},
  journal={ISPRS International Journal of Geo-Information},
  volume={11},
  number={7},
  pages={385},
  year={2022},
  publisher={MDPI}
}

@article{li2024geoai,
  title={GeoAI for Science and the Science of GeoAI},
  author={Li, Wenwen and Arundel, Samantha and Gao, Song and Goodchild, Michael and Hu, Yingjie and Wang, Shaowen and Zipf, Alexander},
  journal={Journal of Spatial Information Science},
  number={29},
  year={2024},
  publisher={Journal of Spatial Information Science}
}

@article{liu2016rethinking,
  title={Rethinking big data: A review on the data quality and usage issues},
  author={Liu, Jianzheng and Li, Jie and Li, Weifeng and Wu, Jiansheng},
  journal={ISPRS journal of photogrammetry and remote sensing},
  volume={115},
  pages={134--142},
  year={2016},
  publisher={Elsevier}
}

@article{li2016geospatial,
  title={Geospatial big data handling theory and methods: A review and research challenges},
  author={Li, Songnian and Dragicevic, Suzana and Castro, Francesc Ant{\'o}n and Sester, Monika and Winter, Stephan and Coltekin, Arzu and Pettit, Christopher and Jiang, Bin and Haworth, James and Stein, Alfred and others},
  journal={ISPRS journal of Photogrammetry and Remote Sensing},
  volume={115},
  pages={119--133},
  year={2016},
  publisher={Elsevier}
}

@article{duckham2000assessment,
  title={Assessment of error in digital vector data using fractal geometry},
  author={Duckham, Matt and Drummond, Jane},
  journal={International Journal of Geographical Information Science},
  volume={14},
  number={1},
  pages={67--84},
  year={2000},
  publisher={Taylor \& Francis}
}

@article{chen2006automatically,
  title={Automatically conflating road vector data with orthoimagery},
  author={Chen, Ching-Chien and Knoblock, Craig A and Shahabi, Cyrus},
  journal={GeoInformatica},
  volume={10},
  number={4},
  pages={495--530},
  year={2006},
  publisher={Springer}
}

@article{mai2023towards,
  title={Towards general-purpose representation learning of polygonal geometries},
  author={Mai, Gengchen and Jiang, Chiyu and Sun, Weiwei and Zhu, Rui and Xuan, Yao and Cai, Ling and Janowicz, Krzysztof and Ermon, Stefano and Lao, Ni},
  journal={GeoInformatica},
  volume={27},
  number={2},
  pages={289--340},
  year={2023},
  publisher={Springer}
}

@article{de2015geographic,
  title={A geographic approach for combining social media and authoritative data towards identifying useful information for disaster management},
  author={De Albuquerque, Joao Porto and Herfort, Benjamin and Brenning, Alexander and Zipf, Alexander},
  journal={International journal of geographical information science},
  volume={29},
  number={4},
  pages={667--689},
  year={2015},
  publisher={Taylor \& Francis}
}

@inproceedings{li2024t,
  title={T-jepa: A joint-embedding predictive architecture for trajectory similarity computation},
  author={Li, Lihuan and Xue, Hao and Song, Yang and Salim, Flora},
  booktitle={Proceedings of the 32nd ACM international conference on advances in geographic information systems},
  pages={569--572},
  year={2024}
}

@inproceedings{zhu2024controltraj,
  title={Controltraj: Controllable trajectory generation with topology-constrained diffusion model},
  author={Zhu, Yuanshao and Yu, James Jianqiao and Zhao, Xiangyu and Liu, Qidong and Ye, Yongchao and Chen, Wei and Zhang, Zijian and Wei, Xuetao and Liang, Yuxuan},
  booktitle={Proceedings of the 30th ACM SIGKDD Conference on Knowledge Discovery and Data Mining},
  pages={4676--4687},
  year={2024}
}

@inproceedings{polygnn,
author = {Yu, Dazhou and Hu, Yuntong and Li, Yun and Zhao, Liang},
title = {PolygonGNN: Representation Learning for Polygonal Geometries with Heterogeneous Visibility Graph},
year = {2024},
isbn = {9798400704901},
publisher = {Association for Computing Machinery},
address = {New York, NY, USA},
url = {https://doi.org/10.1145/3637528.3671738},
doi = {10.1145/3637528.3671738},
booktitle = {Proceedings of the 30th ACM SIGKDD Conference on Knowledge Discovery and Data Mining},
pages = {4012–4022},
numpages = {11},
location = {Barcelona, Spain},
series = {KDD '24}
}

@article{Yan2021,
author = {Xiongfeng Yan and Tinghua Ai and Min Yang and Xiaohua Tong},
title = {Graph convolutional autoencoder model for the shape coding and cognition of buildings in maps},
journal = {International Journal of Geographical Information Science},
volume = {35},
number = {3},
pages = {490--512},
year = {2021},
publisher = {Taylor \& Francis},
doi = {10.1080/13658816.2020.1768260},
URL = {  
        https://doi.org/10.1080/13658816.2020.1768260
},
eprint = {         https://doi.org/10.1080/13658816.2020.1768260
}
}

@inproceedings{oleynikova2016signed,
  title={Signed distance fields: A natural representation for both mapping and planning},
  author={Oleynikova, Helen and Millane, Alexander and Taylor, Zachary and Galceran, Enric and Nieto, Juan and Siegwart, Roland},
  booktitle={RSS 2016 workshop: geometry and beyond-representations, physics, and scene understanding for robotics},
  year={2016},
  organization={University of Michigan}
}

@article{brovelli2017towards,
  title={Towards an automated comparison of OpenStreetMap with authoritative road datasets},
  author={Brovelli, Maria Antonia and Minghini, Marco and Molinari, Monia and Mooney, Peter},
  journal={Transactions in GIS},
  volume={21},
  number={2},
  pages={191--206},
  year={2017},
  publisher={Wiley Online Library}
}

@article{biljecki2023quality,
  title={Quality of crowdsourced geospatial building information: A global assessment of OpenStreetMap attributes},
  author={Biljecki, Filip and Chow, Yoong Shin and Lee, Kay},
  journal={Building and Environment},
  volume={237},
  pages={110295},
  year={2023},
  publisher={Elsevier}
}

\end{document}